\documentclass[runningheads]{llncs}
\usepackage{url}
\usepackage[hidelinks]{hyperref}
\usepackage{graphicx}
\usepackage{amsmath}
\usepackage{amssymb}
\usepackage{booktabs}
\usepackage{algorithm}
\usepackage{algorithmic}
\usepackage[UKenglish]{babel}
\usepackage{microtype}
\usepackage{mathtools}
\usepackage{physics}
\usepackage{tikz}
\usepackage{mathdots}
\usepackage{yhmath}
\usepackage{cancel}
\usepackage{color}
\usepackage{siunitx}
\usepackage{array}
\usepackage{multirow}
\usepackage{gensymb}
\usepackage{tabularx}
\usepackage{extarrows}
\usetikzlibrary{fadings}
\usetikzlibrary{patterns}
\usetikzlibrary{shadows.blur}
\usetikzlibrary{shapes}
\spnewtheorem{principle}{Principle}{\bfseries}{\itshape}

\newcommand{\R}{\mathcal{R}}
\newcommand{\Rs}{\mathcal{R}_s}
\newcommand{\Rd}{\mathcal{R}_d}
\newcommand{\La}{\mathcal{L}}
\newcommand{\K}{\mathcal{K}}
\newcommand{\Kn}{\mathcal{K}_n}
\newcommand{\Kp}{\mathcal{K}_p}
\newcommand{\pr}{^\prime}
\newcommand{\Prem}{\texttt{Prem}}
\newcommand{\Ord}{\texttt{OrdPrem}}
\newcommand{\Ax}{\texttt{Axioms}}
\newcommand{\Conc}{\texttt{Conc}}
\newcommand{\Rules}{\texttt{Rules}}
\newcommand{\DefRules}{\texttt{DefRules}}
\newcommand{\StrRules}{\texttt{StrRules}}
\newcommand{\TopRule}{\texttt{TopRule}}
\newcommand{\Ant}{\texttt{Ant}}
\newcommand{\Sub}{\texttt{Sub}}

\newcommand{\Str}{\texttt{Str}}
\newcommand{\B}{\texttt{Basis}}
\newcommand{\DefB}{\texttt{DefBasis}}
\newcommand{\StrB}{\texttt{StrBasis}}
\newcommand{\Supp}{\texttt{Supp}}
\def\signed #1{{\leavevmode\unskip\nobreak\hfil\penalty50\hskip2em
  \hbox{}\nobreak\hfil[#1]
  \parfillskip=0pt \finalhyphendemerits=0 \endgraf}}
\newsavebox\mybox
\newenvironment{proofStep}[1]
  {\savebox\mybox{#1}}
  {\signed{\usebox\mybox}}
\newcommand{\ProofStep}[2]{\begin{proofStep}{#2}
#1
\end{proofStep}}
\begin{document}
\title{Intrinsic Argument Strength in Structured Argumentation: a Principled Approach}
\titlerunning{Intrinsic Argument Strength in Structured Argumentation}
\author{Jeroen Paul Spaans\orcidID{0000-0001-7027-8102}}
\authorrunning{J.P. Spaans}
\institute{Utrecht University, Utrecht, the Netherlands
\email{jeroen.paul.spaans@gmail.com}}
\maketitle
\begin{abstract}
Abstract argumentation provides us with methods such as gradual and Dung semantics with which to evaluate arguments after potential attacks by other arguments. Some of these methods can take intrinsic strengths of arguments as input, with which to modulate the effects of attacks between arguments. Coming from abstract argumentation, these methods look only at the relations between arguments and not at the structure of the arguments themselves. In structured argumentation the way an argument is constructed, by chaining inference rules starting from premises, is taken into consideration. In this paper we study methods for assigning an argument its intrinsic strength, based on the strengths of the premises and inference rules used to form said argument. We first define a set of principles, which are properties that strength assigning methods might satisfy. We then propose two such methods and analyse which principles they satisfy. Finally, we present a generalised system for creating novel strength assigning methods and speak to the properties of this system regarding the proposed principles.
\keywords{Intrinsic Argument Strength  \and Structured Argumentation \and Principles \and Weight Aggregation \and Aggregation Method.}
\end{abstract}
\section{Introduction}
Argumentation is used in Artificial Intelligence to aid in solving many varied problems; for example, it is used to help with nonmonotonic reasoning \cite{DBLP:journals/ai/Dung95}, to help in making and explaining decisions \cite{DBLP:journals/eswa/ZhongFLT19} and to develop architectures for agents in a multi-agent setting \cite{DBLP:conf/argmas/KakasAKMM11}. Argumentation's core concept is justifying claims by use of arguments. These arguments are reasons to believe or accept a claim.

Arguments might not agree with one another, such as when two arguments support contradicting claims or when one argument contradicts a premise of another. In these cases we speak of one argument standing in an attack relation to another. To help draw conclusions about which arguments to accept  Dung introduced abstract argumentation frameworks \cite{DBLP:journals/ai/Dung95} in which arguments and the binary attack relations between them are modelled in a directed graph. Different semantics may then be used to determine the status of each argument.

Dung's semantics \cite{DBLP:journals/ai/Dung95} and those in the same family, such as those researched in  \cite{DBLP:journals/amai/AmgoudC02}, define extensions of arguments such that every argument is \textit{in} (accepted), \textit{out} (rejected) or, in some cases,  \textit{undecided}.\\
\textit{Gradual semantics}, introduced by Cayrol and Lagasquie-Schiex \cite{DBLP:journals/jair/CayrolL05} and further researched in \cite{DBLP:conf/ijcai/AmgoudBDV17}, do not seek to accept or reject arguments like the aforementioned \textit{Dung-} or \textit{extension semantics} but rather to compute their overall strength.

Arguments will often have a base weight, or \textit{intrinsic strength}, representing, for example, the certainty of the argument's premises \cite{DBLP:conf/uai/BenferhatDP93}. To determine the overall strength of an argument, the attacks against it are taken into account. These attacks may also be weighted, for example to represent their degree of relevance \cite{DBLP:journals/ai/DunneHMPW11}.

Gradual semantics have been proposed both for semi-weighted abstract argumentation frameworks (i.e. those frameworks where only the arguments have a base weight) \cite{DBLP:conf/ijcai/AmgoudBDV17} and for weighted abstract argumentation frameworks (i.e. those frameworks where arguments and attacks are weighted) \cite{DBLP:conf/atal/AmgoudD19}, each based on different ways of aggregating attacks, considering the strength of each attacker and (where applicable) the strength of each attack, to lower the weight or strength of the argument under attack.

The aforementioned falls under what is known as \textit{abstract argumentation}. Here the internal structure of an argument and the nature of attacks is not considered. When we do take these factors into consideration, such as in \cite{DBLP:journals/argcom/ModgilP14,DBLP:conf/kr/Prakken18} and this paper, we speak of \textit{structured argumentation}.

An argument, in structured argumentation, can intuitively be seen as the application of one or more \textit{strict} or \textit{defeasible} inference rules, starting from a set of premises. We might apply the strict inference rule \textit{if X is a bird, then X is an animal} (strict, because this inference is based on a definition and is therefore not open to attack) to the premise \textit{Tweety is a bird} to form an argument for the claim \textit{Tweety is an animal}. Similarly we might apply the defeasible inference rule \textit{if X is a bird, then X can most likely fly} (defeasible because \textit{X} might be a penguin or a baby bird, in which case they cannot fly, leaving the inference open to attack) to the same premise to form an argument for the claim \textit{Tweety can fly}. It is plain to see how inference rules can be combined to form more complex arguments in the shape of an inference tree. If conflict arises between the conclusion of an argument $A$ and some part of an argument $B$ we speak of an attack from $A$ to $B$. we may specify such an attack to, for example, be a \textit{rebuttal} when $A$ attacks $B$ on its conclusion or an \textit{undermining} when it attacks B on a premise. 

While gradual semantics allow us to determine the overall strength of arguments after attacks when we have been given the intrinsic strength of each argument, no standard has arisen in the literature for deriving these intrinsic strengths from the structure of the arguments. Such a standard is what we hope to work toward with this paper. The aim is to investigate different methods for aggregating given weights of the premises and inference rules used to form the argument to derive the intrinsic strength of the argument as a whole.

The paper is structured as follows. We first introduce the basic concepts used in structured argumentation. Then we will define a series of principles, each of which will be a property a method for assigning an argument its intrinsic strength can satisfy. We then propose two intrinsic strength assigning methods and evaluate which principles they satisfy. Next we introduce the aggregation method, a framework for creating new strength assigning methods. Lastly, we speak to the properties of an aggregation method, especially in regard to the earlier-proposed principles.
\section{Basic Concepts}
In argumentation à la Dung, we look at arguments in an argumentation graph, which consists of a set of arguments and a binary attack relation on this set. In gradual argumentation, as in \cite{DBLP:conf/atal/AmgoudD19} which is where we take the following definition from, we often assign weights in the interval $[0,1]$ (lower is weaker) to both the arguments and attacks in our argumentation graph, resulting in a \textit{weighted argumentation graphs}.

\begin{definition}[Weighted Argumentation Graph]\label{WAG}
A weighted argumentation graph (WAG) is an ordered tuple $G = \langle \mathcal{A},\sigma, \mathcal{R},\pi\rangle$, where $\mathcal{A}$ is a non-empty finite set of arguments, $\mathcal{R} \subseteq \mathcal{A} \times \mathcal{A}$, $\sigma : \mathcal{A} \rightarrow [0,1]$ and $\pi : \mathcal{R} \rightarrow [0,1]$.
\end{definition}

Here $\sigma(a)$ is the base weight of argument $a$, $(a,b) \in \mathcal{R}$ means $a$ attacks $b$ and $\pi((a,b))$ is the weight of the attack from $a$ to $b$. In this paper we are looking to formulate $\sigma$ such that it represents an argument's intrinsic strength, where $\sigma(a)$ is an aggregation of the strengths of the premises and inference rules used in $a$ over the structure of $a$.
\begin{example}\label{Ex_Semantics}
Take the argumentation graph in figure \ref{fig:Arg_Graph}. Here $\mathcal{A} = \{a,b,c,d, e\}$ and $\mathcal{R} = \{(a,b),(b,c),(b,e),(d,c)\}$. Assume the weight of all arguments and attacks in the graph is $1$.
\begin{figure}[ht]
    \centering
    \tikzset{every picture/.style={line width=0.75pt}} %set default line width to 0.75pt        
    \begin{tikzpicture}[x=0.75pt,y=0.75pt,yscale=-1,xscale=1]
        %uncomment if require: \path (0,298); %set diagram left start at 0, and has height of 298
        
        %Shape: Circle [id:dp08166983876424505] 
        \draw   (0,175) .. controls (0,161.19) and (11.19,150) .. (25,150) .. controls (38.81,150) and (50,161.19) .. (50,175) .. controls (50,188.81) and (38.81,200) .. (25,200) .. controls (11.19,200) and (0,188.81) .. (0,175) -- cycle ;
        %Shape: Circle [id:dp9132347857537273] 
        \draw   (80,175) .. controls (80,161.19) and (91.19,150) .. (105,150) .. controls (118.81,150) and (130,161.19) .. (130,175) .. controls (130,188.81) and (118.81,200) .. (105,200) .. controls (91.19,200) and (80,188.81) .. (80,175) -- cycle ;
        
        %Curve Lines [id:da5523822242570959] 
        \draw    (130,165) .. controls (139.69,157.79) and (149.2,157.43) .. (158.53,163.92) ;
        \draw [shift={(160,165)}, rotate = 217.77] [color={rgb, 255:red, 0; green, 0; blue, 0 }  ][line width=0.75]    (6.56,-1.97) .. controls (4.17,-0.84) and (1.99,-0.18) .. (0,0) .. controls (1.99,0.18) and (4.17,0.84) .. (6.56,1.97)   ;
        %Curve Lines [id:da220593300232809] 
        \draw    (50,165) .. controls (59.69,157.79) and (69.2,157.43) .. (78.53,163.92) ;
        \draw [shift={(80,165)}, rotate = 217.77] [color={rgb, 255:red, 0; green, 0; blue, 0 }  ][line width=0.75]    (6.56,-1.97) .. controls (4.17,-0.84) and (1.99,-0.18) .. (0,0) .. controls (1.99,0.18) and (4.17,0.84) .. (6.56,1.97)   ;
        %Curve Lines [id:da5239910539451011] 
        \draw    (211.68,163.82) .. controls (221.3,157.43) and (230.74,157.83) .. (240,165) ;
        \draw [shift={(210,165)}, rotate = 323.36] [color={rgb, 255:red, 0; green, 0; blue, 0 }  ][line width=0.75]    (6.56,-1.97) .. controls (4.17,-0.84) and (1.99,-0.18) .. (0,0) .. controls (1.99,0.18) and (4.17,0.84) .. (6.56,1.97)   ;
        %Shape: Circle [id:dp9238327353322894] 
        \draw   (160,175) .. controls (160,161.19) and (171.19,150) .. (185,150) .. controls (198.81,150) and (210,161.19) .. (210,175) .. controls (210,188.81) and (198.81,200) .. (185,200) .. controls (171.19,200) and (160,188.81) .. (160,175) -- cycle ;
        
        %Shape: Circle [id:dp826818548985905] 
        \draw   (240,175) .. controls (240,161.19) and (251.19,150) .. (265,150) .. controls (278.81,150) and (290,161.19) .. (290,175) .. controls (290,188.81) and (278.81,200) .. (265,200) .. controls (251.19,200) and (240,188.81) .. (240,175) -- cycle ;
        
        %Shape: Circle [id:dp814304396026502] 
        \draw   (160,240) .. controls (160,226.19) and (171.19,215) .. (185,215) .. controls (198.81,215) and (210,226.19) .. (210,240) .. controls (210,253.81) and (198.81,265) .. (185,265) .. controls (171.19,265) and (160,253.81) .. (160,240) -- cycle ;
        
        %Curve Lines [id:da6284650887611335] 
        \draw    (115,200) .. controls (120.4,214.93) and (140.52,225.13) .. (158.1,229.54) ;
        \draw [shift={(160,230)}, rotate = 192.93] [color={rgb, 255:red, 0; green, 0; blue, 0 }  ][line width=0.75]    (6.56,-1.97) .. controls (4.17,-0.84) and (1.99,-0.18) .. (0,0) .. controls (1.99,0.18) and (4.17,0.84) .. (6.56,1.97)   ;
        
        % Text Node
        \draw (25,175) node    {$a$};
        % Text Node
        \draw (105,175) node    {$b$};
        % Text Node
        \draw (185,175) node    {$c$};
        % Text Node
        \draw (265,175) node    {$d$};
        % Text Node
        \draw (185,240) node    {$e$};
    \end{tikzpicture}
    \caption{An Argumentation Graph}
    \label{fig:Arg_Graph}
\end{figure}
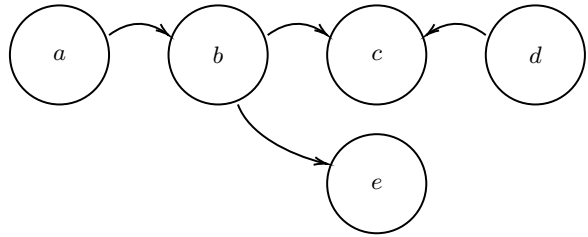

Grounded Dung semantics \cite{DBLP:journals/ai/Dung95} would give us the set of arguments, called an extension, to accept. \textit{Grounded} semantics, specifically, would give the smallest complete extension.
To find this extension we need a few concepts:
\begin{itemize}
    \item an argument $a \in \mathcal{A}$ is acceptable with respect to $E\subseteq \mathcal{A}$ iff $E$ defends $a$. That is, $\forall b \in \mathcal{A}$ s.t. $(b,a)\in \mathcal{R}$,$\exists c \in E$ s.t. $(c,b) \in \mathcal{R}$.
    \item A set of arguments $E$ is conflict free iff $\forall a,b \in E, (a,b) \notin \mathcal{R}$.
    \item A set of arguments $E$ is admissible iff it is conflict-free and all arguments in $E$ are acceptable with respect to $E$.
    \item A set of arguments $E$ is a complete extension iff it is an admissible set and every acceptable argument with respect to $E$ belongs to $E$.
\end{itemize}
Following these concepts we see that $b$ and $c$ cannot be included in a complete extension, since $a$ and $d$ have no attackers and as such a set containing $b$ and $c$ could never defend them from the attacks of $a$ and $d$. From here we get the unique grounded extension $E_g = \{a,d,e\}$.

The gradual Weighted h-Categorizer Semantics \cite{DBLP:conf/ijcai/AmgoudBDV17} (which we can use because all attack weights are $1$) would assign each argument $x$ an acceptability degree $\text{Deg}(x)=\lim_{i \rightarrow \infty}f^i(x)$ where $$f^i(x) = \begin{cases} \sigma(x) &\mbox{if } i=0 \\
\frac{\sigma(x)}{1+\sum_{b_i\in \texttt{Att}(x)}f^{i-1}(b_i)} &\mbox{otherwise }\end{cases}$$ and $\texttt{Att}(x)$ denotes the attackers of $x$. This would result in $\text{Deg}(a)=1$, $\text{Deg}(b)=\frac{1}{2}$, $\text{Deg}(c)=\frac{2}{5}$, $\text{Deg}(d)=1$ and $\text{Deg}(e)=\frac{2}{3}$. 
\end{example}

Having seen how arguments can relate to one another, we now look to how arguments are formed. In doing so, we introduce a modified variant of the ASPIC+ framework \cite{DBLP:journals/argcom/ModgilP14}.

To construct an argument, we must first know the building blocks that are at our disposal. In an argument we make inferences based on inference rules with antecedents and consequents that are all well-formed formulae in some logical language.

\begin{definition}[Argumentation System]\label{AS}
An argumentation system is a pair $\text{AS} = ( \La,\R )$ where:
\begin{itemize}
    \item $\La$ is a logical language consisting of propositional or ground predicate-logic literals that is closed under negation.
    \item $\R = \Rd \cup \Rs$ with $\Rd \cap \Rs = \emptyset$, where $\Rd$ is a finite set of defeasible inference rules of the form $\{\varphi_1, \dots,\varphi_n\} \Rightarrow \varphi$, $\Rs$ is a finite set of strict inference rules of the form $\{\varphi_1, \dots,\varphi_n\} \rightarrow \varphi$ and $\varphi, \varphi_i$ are meta-variables ranging over well-formed formulae in $\La$. We call $\varphi_1, \dots,\varphi_n$%\footnote{Below, we will omit the brackets around the antecedents.}
    the antecedents of the rule and $\varphi$ its consequent.
\end{itemize}
\end{definition}

Just in case $\psi = \neg \varphi$ or $\varphi = \neg \psi$, we write $\psi = -\varphi$. Here $-$ is not a member of $\La$ but rather a metalinguistic symbol used to simplify notation.

In any argument we start our reasoning from one or more \textit{premises}. These are the pieces of knowledge from which we infer other information. What is important to notice, is that a premise may be \textit{fallible} or \textit{infallible}. Fallible premises are open to attack. Suppose we combine our belief that we saw Robin in Rotterdam this morning with the knowledge that Rotterdam is in the Netherlands to argue Robin was in the Netherlands this morning. This argument is deductively valid but still open to attack. Suppose our friend Alex informs us they saw Robin in Berlin at the same time we believe to have seen them in Rotterdam. Since our friends usually tell us the truth, this allows us to form an argument that attacks the original premise that we saw Robin in Rotterdam. Infallible premises, such as \textit{1 is a natural number}, are not open to attack. In accordance with \cite{DBLP:journals/argcom/ModgilP14} we will call these infallible premises \textit{axiom} premises and we will refer to premises that are open to attack as \textit{ordinary} premises. We call the body of information from which premises may be taken a knowledge base.

\begin{definition}[Consistency]\label{Consistency}
For any $S \subseteq \La$, let the closure of $S$ under strict rules, denoted $\text{Cl}_{\mathcal{R}_{S}}(S)$, be the smallest set containing S and the consequent of any strict rule in $\Rs$ whose antecedents are in $\text{Cl}_{\mathcal{R}_{S}}(S)$. Then a set $S \subseteq \La$ is directly consistent iff there are no $\psi, \varphi \in S$ such that $\psi = -\varphi$ and indirectly consistent iff $\text{Cl}_{\mathcal{R}_{S}}(S)$ is directly consistent. \cite{DBLP:conf/kr/Prakken18}
\end{definition}

\begin{definition}[Knowledge Base]\label{KB}
A knowledge base in an $\text{AS} = ( \La,\R )$ is a set $\K \subseteq \La$, where $\K = \Kn \cup \Kp$, $\Kn$ is a set of axioms, $\Kp$ is a set of ordinary premises, $\Kn$ is indirectly consistent and $\Kn \cap \Kp = \emptyset$.% \comment{Maybe make ordinary premises directly consistent.}
\end{definition}

With an argumentation system and a knowledge base we could create an argument, but we would still be missing the rule and premise weights. To codify these we introduce the \textit{weighted argumentation theory}.

\begin{definition}[Weighted Argumentation Theory]\label{WAT}
A weighted argumentation theory is a tuple $\text{WAT} = ( \text{AS}, \K, s )$ where:
\begin{itemize}
    \item $\text{AS} = ( \La,\R )$ is an argumentation system. %\comment{Maybe disallow inferring contradictory consequents from the same antecedents.}
    \item $\K$ is a knowledge base.
    \item $s$ is a function assigning weights to rules and premises, such that $\forall r \in \Rs,\ s(r) = 1;\ \forall p \in \Kn,\ s(p) = 1;\ \forall r\pr \in \Rd,\ s(r\pr) \in [0,1);\ \forall p\pr \in \Kp,\ s(p\pr) \in [0,1)$ and a higher weight is assigned to stronger premises and inference rules.
\end{itemize}
\end{definition}

\begin{example}
Continuing with the \textit{Tweety is a bird} example from the introduction to this paper, we might have:
\begin{itemize}
    \item $s(\text{Tweety is a bird} \rightarrow \text{Tweety is an animal}) = 1$; because all birds are, by definition, animals and as such this is a strict inference rule.
    \item $s(\text{Tweety is a bird} \Rightarrow \text{Tweety can fly}) = 0.95$; because most birds can fly, so this is a strong defeasible inference rule.
    \item $s(\text{Tweety is a bird} \Rightarrow \text{Tweety is yellow}) = 0.05$; because, while existent, yellow birds are quite rare, so this is a weak defeasible inference rule.
\end{itemize}
\end{example}

We can now define an argument over a $\text{WAT} = ( \text{AS}, \K, s )$. As in ASPIC+ \cite{DBLP:journals/argcom/ModgilP14} we chain together applications of inference rules from $\text{AS}$ into inference trees, starting from premises in $\K$.

For a given argument $A$, $\Conc(A)$ returns the conclusion of $A$, $\TopRule(A)$ returns the last inference rule used in the argument, $\Ant(A)$ returns the argument's set of antecedent arguments, $\Sub(A)$ returns the subarguments of $A$, $\DefRules(A)$ returns all the defeasible rules used in $A$, $\StrRules(A)$ returns all the used strict rules and $\Ord(A)$ and $\Ax(A)$ return the ordinary and axiomatic premises used to construct the argument respectively.

The structure of an argument defined like this is the same as in ASPIC+, but there are two differences between the functions we define over an argument and those commonly used in ASPIC+. Firstly, we split up the function $\Prem$, which returns all of the premises used in an argument,  into $\Ax$ and $\Ord$ to more easily distinguish between the strict and defeasible parts of an argument. Secondly and more notably, while ASPIC+ uses ordinary sets for the values of $\Prem$, $\StrRules$ and $\DefRules$, we use multisets for the values of the four premise and rule functions. A multiset, also called a bag, is much like an ordinary set but, contrary to a normal set, is able to contain an element more than once \cite{DBLP:books/lib/Knuth98}. The use of multisets allows us to more easily determine the strength of an argument in later sections of this paper because, intuitively, each use of a defeasible premise or inference rule should affect the strength of an argument and these multisets allow us to easily iterate over each occurence.

\begin{definition}[Multiset]
A multiset is a modification of a set that allows multiple instances of its elements. Like an ordinary set, a multiset is unordered.
e.g. $[a,a,a,b,b]$ is a multiset containing $a$ and $b$ where $a$ has multiplicity $m(a) = 3$ and $b$ has multiplicity $m(b) = 2$. This multiset may also be denoted $[a^3,b^2]$ or $\{(a,3),(b,2)\}$.

We often say the elements of a multiset come from a fixed set $U$ called the universe, such that the support of a multiset $A$ is the multiset's underlying set $\Supp(A) = \{x \in U | m_A(x) > 0\}$. For readability, we say $A=\emptyset$ when $\Supp(A) = \emptyset$.

We use the following functions on multisets:
\begin{itemize}
\item Union: the union of multisets $A$ and $B$, $A\cup B$ is the multiset $C$ with multiplicity function $m_C(x) = \text{max}(m_A(x),m_B(x)), \forall x \in U$.
\item Sum: the sum of of multisets $A$ and $B$, $A\uplus B$ is the multiset $C$ with multiplicity function $m_C(x) = m_A(x) + m_B(x), \forall x \in U$.
% \item Membership: for a multiset $A$, when we use set membership as in $\prod_{a \in A}f(a)$, $\{a \in A | p(a)\}$ or $\forall a \in A, q(a)$ we take this to be based on individual occurrences of an element. e.g. $\prod_{a \in [b,b,c]}a = b^2c$.
\end{itemize}
When using multisets in product or sum notation we assume to iterate over each occurrence of an element; for instance $\prod_{a \in [b,b,c]}a = b \cdot b \cdot c$. The same assumption is made for set builder notation, such that $\{f(x) | x \in [a,a,b]\} = [f(a),f(a),f(b)]$.
\end{definition}

\begin{definition}[(General) Argument]\label{Argument}
A general argument  $A$ over a $\text{WAT} = ( \text{AS}, \K, s )$ is defined recursively. It can be obtained by applying one or more of the following steps a finite amount of times;
\begin{enumerate}
    \item premise $\varphi$, if $\varphi \in \Kn$, where:\\
    $\Conc(A) = \varphi$;\\
    $\TopRule(A) = undefined$;\\
    $\Ant(A) = \emptyset$;\\
    $\Sub(A) = \{\varphi\}$;\\
    $\DefRules(A) = \emptyset$;\\
    $\StrRules(A) = \emptyset$;\\
    $\Ord(A) = \emptyset$;\\
    $\Ax(A) = [\varphi]$.
    \item premise $\varphi$, if $\varphi \in \Kp$, where:\\
    $\Conc(A) = \varphi$;\\
    $\TopRule(A) = undefined$;\\
    $\Ant(A) = \emptyset$;\\
    $\Sub(A) = \{\varphi\}$;\\
    $\DefRules(A) = \emptyset$;\\
    $\StrRules(A) = \emptyset$;\\
    $\Ord(A) = [\varphi]$;\\
    $\Ax(A) = \emptyset$.
    \item $\{A_1, \dots, A_n\} \rightarrow \varphi$, if $A_1, \dots, A_n$ are arguments and \\$\{\Conc(A_1), \dots, \Conc(A_n)\} \rightarrow \varphi \in \Rs$, where:\\
    $\Conc(A) = \varphi$;\\
    $\TopRule(A) = \{\Conc(A_1), \dots, \Conc(A_n)\} \rightarrow \varphi$;\\
    $\Ant(A) = \{ A_1, \dots, A_n \}$;\\
    $\Sub(A) = \Sub(A_1) \cup \dots \cup \Sub(A_n) \cup \{A\}$;\\
    $\DefRules(A) = \DefRules(A_1) \uplus \dots \uplus \DefRules(A_n)$;\\
    $\StrRules(A) \\= \StrRules(A_1) \uplus \dots \uplus \StrRules(A_n) \uplus [\{\Conc(A_1), \dots, \Conc(A_n)\} \rightarrow \varphi]$;\\
    $\Ord(A) = \Ord(A_1) \uplus \dots \uplus \Ord(A_n)$;\\
    $\Ax(A) = \Ax(A_1) \uplus \dots \uplus \Ax(A_n)$
    \item $\{A_1, \dots, A_n\} \Rightarrow \varphi$, if $A_1, \dots, A_n$ are arguments and \\$\{\Conc(A_1), \dots, \Conc(A_n)\} \Rightarrow \varphi \in \Rd$, where:\\
    $\Conc(A) = \varphi$;\\
    $\TopRule(A) = \{\Conc(A_1), \dots, \Conc(A_n)\} \Rightarrow \varphi$;\\
    $\Ant(A) = \{ A_1, \dots, A_n \}$;\\
    $\Sub(A) = \Sub(A_1) \cup \dots \cup \Sub(A_n) \cup \{A\}$;\\
    $\DefRules(A) \\= \DefRules(A_1) \uplus \dots \uplus \DefRules(A_n) \uplus [\{\Conc(A_1), \dots, \Conc(A_n)\} \Rightarrow \varphi]$;\\
    $\StrRules(A) = \StrRules(A_1) \uplus \dots \uplus \StrRules(A_n)$;\\
    $\Ord(A) = \Ord(A_1) \uplus \dots \uplus \Ord(A_n)$;\\
    $\Ax(A) = \Ax(A_1) \uplus \dots \uplus \Ax(A_n)$.
\end{enumerate}

A general argument $A$ is an argument iff:
\begin{enumerate}
    \item $\Sub(A)$ is indirectly consistent; and
    \item If $A$ contains non-strict subarguments $A\pr$ and $A^{\prime\prime}$ such that $\Conc(A\pr) = \Conc(A^{\prime\prime})$, then $A\pr = A^{\prime\prime}$.
\end{enumerate}
An argument $A$ is called strict when $\DefRules(A) = \Ord(A) = \emptyset$. Else it is called defeasible.
\end{definition}

Later in this paper, we will use the premises and inference rules used in an argument as the input of functions to determine the argument's strength. To simplify notation we introduce the \textit{basis} of an argument.

\begin{definition}[(Defeasible/Strict) Basis]\label{Basis}
For an argument $A$ the defeasible basis of $A$, written $\DefB(A)$ is the multiset of all ordinary premises and defeasible inference rules used in $A$. i.e. $\DefB(A) = \Ord(A) \uplus \DefRules(A)$.

The strict basis of $A$, $\StrB(A) = \Ax(A) \uplus \StrRules(A)$, is the multiset of all axiomatic premises and strict inference rules used in $A$.

The basis of $A$ is the sum of its strict and defeasible bases, s.t. $\B(A) = \DefB \uplus \StrB$.
\end{definition}

For ease of notation we also introduce the $\Rules$ and $\Prem$ functions which return the multisets of all inference rules and premises used in an argument respectively.

\begin{definition}[$\Rules$]\label{Rules}
For an argument $A$, \begin{gather*}
    \Rules(A) = \StrRules(A) \uplus \DefRules(A).
\end{gather*}
\end{definition}
\begin{definition}[$\Prem$]\label{Prem}
For an argument $A$, \begin{gather*}
    \Prem(A) = \Ax(A) \uplus \Ord(A).
\end{gather*}
\end{definition}

We say two arguments are isomorphic if they have the same structure (or shape) and have the same weights for the equivalently positioned premises and inference rules.
\begin{definition}[Isomorphism] \label{Isomorphism}
Take arguments $A$ over $\text{WAT} = ( \text{AS}, \K, s )$ and $A\pr$ over a $\text{WAT}\pr = ( \text{AS}\pr, \K\pr, s\pr )$. There exists an isomorphism between $A$ and $A\pr$ when:
\begin{itemize}
    \item If the arguments state a premise (Items 1, 2; Definition \ref{Argument}),
    \begin{gather*}
        s(\Conc(A)) = s\pr(\Conc(A\pr)).
    \end{gather*}
    \item If the arguments make an inference (Items 3, 4; Definition \ref{Argument}),
    \begin{multline*}
        s(\TopRule(A)) = s\pr(\TopRule(A\pr))\text{ and} \\\text{there exists a bijective function }f:\Ant(A) \rightarrow \Ant({A\pr}) \text{ such that }\\\forall A^{\prime\prime} \in \Ant(A),f(A^{\prime\prime}) \in \Ant(A\pr) \text{ is an isomorphic image of } A^{\prime\prime}.
    \end{multline*}
\end{itemize}
\end{definition}

With our arguments defined, we are looking to assign each argument an intrinsic strength, based on its structure.

\begin{definition}[Intrinsic Strength] \label{Strength}
$\Str$ is a function that assigns numbers in $[0,1]$ to arguments, such that for an argument $A$ over a $\text{WAT} = ( \text{AS}, \K, s )$, $\Str(A)$ is the intrinsic strength of $A$ where stronger arguments are assigned higher values.
\end{definition}
\section{Principles}\label{Principles}

When assigning intrinsic strength to arguments, we may wish to look to certain principles our method of assigning these strengths might adhere to. This aids us in understanding the method we use to assign strengths, in comparing different strength-assigning methods, in proposing sensible methods for assigning strengths, and in selecting a suitable method for assigning strength for a certain application.

In this section we propose 13 such principles, which describe the way the weights of premises and inference rules and the intrinsic strengths of antecedent arguments affect the intrinsic strength of an argument. Most of these principles are intended to be intuitively desirable traits for the assigning of a strength value to an argument.

A similar approach is taken in \cite{DBLP:conf/ijcai/AmgoudBDV17} and later in \cite{DBLP:conf/atal/AmgoudD19} in the exploration of semantics that assign acceptability degrees to arguments after attacks by other arguments in (semi-)weighted argumentation graphs by aggregating attacks, (their weights,~) and the base argument weights. Many of the principles proposed in this section resemble those used in the aforementioned papers.

Our first principle states that the identity, that is the name and meaning, of an argument $A$ should not affect the strength assigned to it. Only its structure should.
\begin{principle}[Anonymity]\label{Anonymity}
\begin{multline*}
    \forall \text{WAT} = ( \text{AS}, \K, s ),\ \forall A\, A\pr \text{ over WAT},\\\text{ if an isomorphism exists between } A \text{ and } A\pr,\  \Str(A) = \Str(A\pr)
\end{multline*}
\end{principle}

The second principle says that when an argument $A$ only states a premise, the argument should have a strength equal to the weight of its premise.
\begin{principle}[Premising]
\begin{multline*}
    \forall \text{WAT} = ( \text{AS}, \K, s ),\ \forall A \text{ over WAT},\\TopRule(A) = \text{ undefined } \rightarrow \Str(A) = s(\Conc(A))
\end{multline*}
\end{principle}

Our next principle prescribes that when all of an argument $A$'s premises are certain and its inferences are strict, the argument's strength should be 1.
\begin{principle}[Strict Argument]\leavevmode
\begin{gather*}
    \forall \text{WAT} = ( \text{AS}, \K, s ),\ \forall A \text{ over WAT},\ \DefB(A) = \emptyset \rightarrow \Str(A) = 1
\end{gather*}
\end{principle}

The Resilience principle states that when all premises and inference rules used in an argument have a weight higher than 0, the argument's strength should also be higher than 0.
\begin{principle}[Resilience]
\begin{gather*}
    \forall \text{WAT} = ( \text{AS}, \K, s ),\ \forall A \text{ over WAT},\ (\forall b \in \B(A),\ s(b) > 0) \rightarrow \Str(A) > 0
\end{gather*}
\end{principle}

According to the next principle, if any of the premises or inference rules used in argument $A$ has weight 0, the intrinsic strength of $A$ should also be 0.
\begin{principle}[Argument Death]
\begin{gather*}
    \forall \text{WAT} = ( \text{AS}, \K, s ),\ \forall A \text{ over WAT},\ (\exists b \in \B(A),\ s(b) = 0) \rightarrow \Str(A) = 0
\end{gather*}
\end{principle}

% Inference Maximality states that when an argument makes a strict inference, its strength should be no lower than that of its weakest antecedent.
% \begin{principle}[Inference Maximality]
% $\forall \text{WAT} = ( \text{AS}, \K, s ) \forall A \text{ over WAT }, s(TopRule(A)) = 1 \rightarrow \Str(A) \nless \text{min}\{ \Str(A\pr) | A\pr \in \Ant(A)\}$
% \end{principle}

The next principle says that when all antecedents of an argument $A$ have intrinsic strength 1, the intrinsic strength of $A$ should equal the weight of its top rule.
\begin{principle}[Antecedent Maximality]
\begin{multline*}
    \forall \text{WAT} = ( \text{AS}, \K, s ),\ \forall A \text{ over WAT},\ \forall A\pr \in \Ant(A),\\\Str(A\pr) = 1 \wedge \TopRule(A) \neq \text{ undefined } \rightarrow \Str(A) = s(\TopRule(A))
\end{multline*}
\end{principle}

Next, Antecedent Neutrality says that any antecedents of an argument $A$ with intrinsic strength 1 should not affect the intrinsic strength of $A$.
\begin{principle}[Antecedent Neutrality]
\begin{multline*}
    \forall \text{WAT} = ( \text{AS}, \K, s ),\ \forall A\, A\pr\, A^{\prime\prime} \text{ over WAT},\\s(\TopRule(A)) = s(\TopRule(A\pr)) \wedge \Ant(A\pr) = \Ant(A) \cup \{A^{\prime\prime}\} \wedge \Str(A^{\prime\prime}) = 1\\\rightarrow \Str(A) = \Str(A\pr)
\end{multline*}
\end{principle}

Dual to Antecedent Neutrality, Antecedent Weakening says that any antecedents of an argument $A$ with intrinsic strength lower than 1 should lower the intrinsic strength of $A$ if it is not already 0.
\begin{principle}[Antecedent Weakening]
\begin{multline*}
    \forall \text{WAT} = ( \text{AS}, \K, s ),\ \forall A\, A\pr\, A^{\prime\prime} \text{ over WAT},\\s(\TopRule(A)) = s(\TopRule(A\pr)) \wedge \Ant(A\pr) = \Ant(A) \cup \{A^{\prime\prime}\}\\\wedge \Str(A^{\prime\prime}) < 1 \wedge \Str(A) > 0\\\rightarrow \Str(A) > \Str(A\pr)
\end{multline*}
\end{principle}

To satisfy Inferential Weakening, applying a defeasible inference rule should result in an argument with an intrinsic strength lower than that of any of its antecedent arguments, so long as none of the antecedent arguments have strength 0.
\begin{principle}[Inferential Weakening]
\begin{multline*}
    \forall \text{WAT} = ( \text{AS}, \K, s ),\ \forall A \text{ over WAT},\\TopRule(A) \in \Rd \wedge (\forall A\pr \in \Ant(A), Str(A\pr) > 0)\\\rightarrow \Str(A) < \text{min}\{ \Str(A\pr) | A\pr \in \Ant(A)\}
\end{multline*}
\end{principle}

For Inference Weight Sensitivity to apply, applying a weaker defeasible inference rule to a set of antecedent arguments should result in a weaker argument than applying a stronger rule to the same antecedents would, so long as none of the antecedent arguments have strength 0.
\begin{principle}[Inference Weight Sensitivity]
\begin{multline*}
    \forall \text{WAT} = ( \text{AS}, \K, s ),\ \forall A\, A\pr \text{ over WAT},\\\Ant(A) = \Ant(A\pr) \wedge s(\TopRule(A)) < s(\TopRule(A\pr))\\\wedge (\forall A^{\prime\prime} \in \Ant(A),\ \Str(A^{\prime\prime}) > 0)\\\rightarrow \Str(A) < \Str(A\pr)
\end{multline*}
\end{principle}

The Proportionality principle says that when two arguments have equally strong top rules and for each of the first argument's antecedents the second argument has a distinct antecedent with a lower strength, the first argument's overall strength should be higher.
\begin{principle}[Proportionality]
\begin{multline*}
    \forall \text{WAT} = ( \text{AS}, \K, s ),\ \forall A\, A\pr \text{ over WAT},\\s(\TopRule(A)) = s(\TopRule(A\pr)) \\\wedge \text{ there exists an injective function } f: \Ant(A) \rightarrow \Ant(A\pr) \\\text{such that } \forall A^{\prime\prime} \in \Ant(A), \Str(A^{\prime\prime}) > \Str(f(A^{\prime\prime}))\\\rightarrow \Str(A) > \Str(A\pr)
\end{multline*}
\end{principle}

% For Counting to be satisfied, when two arguments have equally strong top rules, but one has more non-strict antecedents, that arguments overall strength should be lower.
% \begin{principle}[Counting]
% \begin{multline*}
%     \forall \text{WAT} = ( \text{AS}, \K, s ) \forall A A\pr \text{ over WAT},\\s(\TopRule(A)) = s(\TopRule(A\pr))\\\wedge |\{ a \in \Ant(A)|\Str(a) < 1\}| < |\{ a\pr \in \Ant(A\pr)|\Str(a\pr) < 1\}|\\\rightarrow \Str(A) > \Str(A\pr)
% \end{multline*}
% \end{principle}

Prescribing a single correct valuation, Weakest Link says an argument $A$'s strength should be equal to the weight of its weakest premise or inference rule.
\begin{principle}[Weakest Link]
\begin{gather*}
    \forall \text{WAT} = ( \text{AS}, \K, s ),\ \forall A \text{ over WAT},\ \Str(A) = \text{min}(\{s(b)| b\in \B(A)\})
\end{gather*}
\end{principle}

The Weakest Link principle is so restrictive in what strengths it allows that most methods of assigning strength will not satisfy it. Still, limiting the strength of an argument to the weight of its weakest link seems to be a desirable property. To accommodate this we introduce the Weakest-Link Limiting principle, which states the intrinsic strength should be no higher than the weight of its weakest premise or inference rule.
\begin{principle}[Weakest-Link Limiting]
\begin{gather*}
    \forall \text{WAT} = ( \text{AS}, \K, s ),\ \forall A \text{ over WAT},\ \Str(A) \leq \text{min}(\{s(b)| b\in \B(A)\})
\end{gather*}
\end{principle}
\section{Assigning Intrinsic Strength}\label{Assigning_Strength}
Having defined a series of principles by which to evaluate methods for assigning intrinsic strength to arguments, we now look to what said methods might be. Two methods that come to mind are the \textit{simple product} method, where we multiply the weights of all the premises and inference rules used in an argument (adding a factor for each time a premise or rule is used) to determine this arguments strength, and the \textit{weakest link} method, where we equate the intrinsic strength of an argument to the lowest weight of any premise or inference rule used in it.

\begin{definition}[Simple Product Method]\label{SP}
The simple product method (SP) assigns any argument $A \text{ over any WAT } = ( \text{AS}, \K, s )$ an intrinsic strength equal to the product of the weights of all members of $\B(A)$, such that: $$\Str_{sp}(A) = \prod_{b\in\B(A)}s(b)$$
Because $\B(A) = \DefB(A) \uplus \StrB(A)$,  $\forall b \in \StrB(A), s(b) = 1$ (Definitions \ref{WAT}, \ref{Argument} and \ref{Basis}) and $1$ is the identity element for multiplication, this is equivalent to:
$$\Str_{sp}(A) = \prod_{b\in\DefB(A)}s(b)$$
\end{definition}

\begin{definition}[Weakest Link Method]\label{WL}
The weakest link method (WL) assigns any argument $A \text{ over any WAT } = ( \text{AS}, \K, s )$ an intrinsic strength equal to the minimum of the weights of all members of $\B(A)$, such that: $$\Str_{wl}(A) = \text{min}\{s(b)|b\in\B(A)\}$$
\end{definition}

\begin{example}\label{Example_Argument}
We look at an example argument $A_4$ with subarguments $A_1$, $A_2$ and $A_3$, shown in figure \ref{fig:argument}. Argument $A_1$ states an axiomatic premise with weight 1. $A_2$ makes an inference from the premise stated in $A_1$ using defeasible inference rule $d_1$ with strength $\frac{1}{4}$. Argument $A_3$ states an ordinary premise with weight $\frac{1}{2}$. Finally, argument $A_4$ uses strict inference rule $s_1$ to infer its conclusion from the conclusion of $A_2$ and the premise stated in $A_3$.
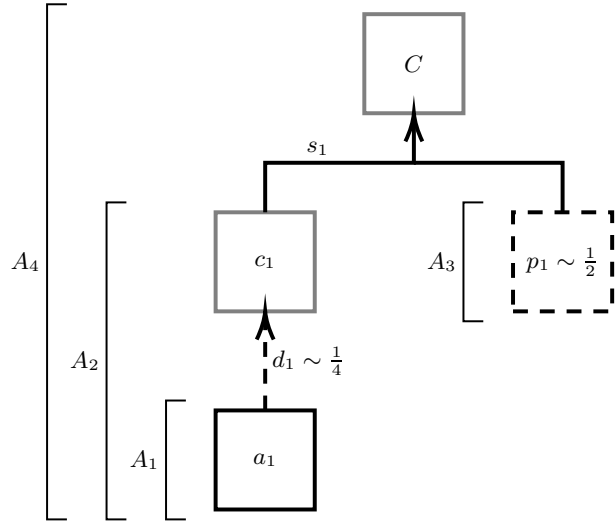
\begin{figure}[ht]
    \centering
    \tikzset{every picture/.style={line width=0.75pt}} %set default line width to 0.75pt        
    
    \begin{tikzpicture}[x=0.75pt,y=0.75pt,yscale=-1,xscale=1]
%uncomment if require: \path (0,505); %set diagram left start at 0, and has height of 505

%Shape: Square [id:dp2663696209067795] 
\draw  [dash pattern={on 5.63pt off 4.5pt}][line width=1.5]  (375,250) -- (425,250) -- (425,300) -- (375,300) -- cycle ;

%Shape: Square [id:dp017519806086913836] 
\draw  [color={rgb, 255:red, 128; green, 128; blue, 128 }  ,draw opacity=1 ][line width=1.5]  (225,250) -- (275,250) -- (275,300) -- (225,300) -- cycle ;
%Shape: Square [id:dp015629567316142734] 
\draw  [color={rgb, 255:red, 128; green, 128; blue, 128 }  ,draw opacity=1 ][line width=1.5]  (300,150) -- (350,150) -- (350,200) -- (300,200) -- cycle ;
%Shape: Square [id:dp32042611346286953] 
\draw  [line width=1.5]  (225,350) -- (275,350) -- (275,400) -- (225,400) -- cycle ;

%Straight Lines [id:da6304261407576469] 
\draw [line width=1.5]  [dash pattern={on 5.63pt off 4.5pt}]  (250,350) -- (250,303) ;
\draw [shift={(250,300)}, rotate = 450] [color={rgb, 255:red, 0; green, 0; blue, 0 }  ][line width=1.5]    (14.21,-4.28) .. controls (9.04,-1.82) and (4.3,-0.39) .. (0,0) .. controls (4.3,0.39) and (9.04,1.82) .. (14.21,4.28)   ;
%Straight Lines [id:da5139796216610978] 
\draw [line width=1.5]    (250,250) -- (250,225) -- (325,225) -- (325,203) ;
\draw [shift={(325,200)}, rotate = 450] [color={rgb, 255:red, 0; green, 0; blue, 0 }  ][line width=1.5]    (14.21,-4.28) .. controls (9.04,-1.82) and (4.3,-0.39) .. (0,0) .. controls (4.3,0.39) and (9.04,1.82) .. (14.21,4.28)   ;
%Straight Lines [id:da2569172055421678] 
\draw [line width=1.5]    (325,225) -- (400,225) -- (400,250) ;
%Straight Lines [id:da45603356336146184] 
\draw    (210,345) -- (200,345) -- (200,375) -- (200,405) -- (210,405) ;
%Straight Lines [id:da9182655103661508] 
\draw    (360,245) -- (350,245) -- (350,275) -- (350,305) -- (360,305) ;
%Straight Lines [id:da8094208164445741] 
\draw    (180,245) -- (170,245) -- (170,325) -- (170,405) -- (180,405) ;

%Straight Lines [id:da19024831671456854] 
\draw    (150,145) -- (140,145) -- (140,275) -- (140,405) -- (150,405) ;

% Text Node
\draw (250,375) node    {$a_{1}$};
% Text Node
\draw (400,275) node    {$p_{1} \sim \frac{1}{2}$};
% Text Node
\draw (252,325) node [anchor=west] [inner sep=0.75pt]    {$d_{1} \sim \frac{1}{4}$};
% Text Node
\draw (276.5,222.6) node [anchor=south] [inner sep=0.75pt]    {$s_{1}$};
% Text Node
\draw (325,175) node    {$C$};
% Text Node
\draw (250,275) node    {$c_{1}$};
% Text Node
\draw (198,375) node [anchor=east] [inner sep=0.75pt]    {$A_{1}$};
% Text Node
\draw (348,275) node [anchor=east] [inner sep=0.75pt]    {$A_{3}$};
% Text Node
\draw (138,275) node [anchor=east] [inner sep=0.75pt]    {$A_{4}$};
% Text Node
\draw (168,325) node [anchor=east] [inner sep=0.75pt]    {$A_{2}$};

\end{tikzpicture}
    \caption{An Argument}
    \label{fig:argument}
\end{figure}

If we were to use the simple product method to assign strengths to $A_4$ and its subarguments, we would have:
\begin{itemize}
    \item $\Str_{sp}(A_1) = 1$ as it uses no rules and states an axiomatic premise.
    \item $\Str_{sp}(A_2) = 1 \cdot \frac{1}{4} = \frac{1}{4}$; multiplying the weight of its one premise and its one defeasible inference rule.
    \item $\Str_{sp}(A_3) = \frac{1}{2}$; the weight of the ordinary premise it states.
    \item $\Str_{sp}(A_4) = 1 \cdot 1 \cdot \frac{1}{4} \cdot \frac{1}{2} = \frac{1}{8}$; the product of the weights of all the inference rules and premises used in the argument.
\end{itemize}

If we were to use the weakest link method instead, we would have:
\begin{itemize}
    \item $\Str_{wl}(A_1) = 1$ as it uses no rules and states an axiomatic premise.
    \item $\Str_{wl}(A_2) = \text{min}(1,\frac{1}{4}) = \frac{1}{4}$; the smallest weight of its one premise and its one defeasible inference rule.
    \item $\Str_{wl}(A_3) = \frac{1}{2}$; the weight of the ordinary premise it states.
    \item $\Str_{wl}(A_4) = \text{min}(1, \frac{1}{4}, \frac{1}{2}) = \frac{1}{4}$; the smallest of the weights of all the inference rules and premises used in the argument.
\end{itemize}
\end{example}

Note that for SP, because of the commutative and associative properties of multiplication, instead of assigning strength equal to the product of the weights of all the premises and inference rules used in an argument, we can equivalently assign the weight of the premise for arguments just stating a premise and multiply the weight of the inference rule used with the strengths of the antecedent arguments for an argument making an inference. This method better matches the recursive nature of an argument. Similarly, we can keep the weight of the premise for arguments just stating a premise and take the minimum of the weight of the inference rule used and the strengths of the antecedent arguments for an argument making an inference when using WL.

\begin{theorem}[SP Works Recursively]\label{T_SP_Rec}
For an argument $A$ over a $\text{WAT } = ( \text{AS}, \K, s )$:
\begin{itemize}
    \item if $\TopRule(A) = \text{ undefined}$, $\Str_{sp}(A) = s(\Conc(A))$
    \item else, $\Str_{sp}(A) = s(\TopRule(A)) \cdot \prod_{a \in \Ant(A)}\Str_{sp}(a)$
\end{itemize}
\end{theorem}

\begin{theorem}[WL Works Recursively]\label{T_WL_Rec}
For an argument $A$ over a $\text{WAT } = ( \text{AS}, \K, s )$:
\begin{itemize}
    \item if $\TopRule(A) = \text{ undefined}$, $\Str_{wl}(A) = s(\Conc(A))$
    \item else, $\Str_{wl}(A) = \text{min}(s(\TopRule(A)), m)$ \\where $m = \text{min}(\{\Str_{wl}(a)|a \in \Ant(A)\})$
\end{itemize}
\end{theorem}
\begin{note}
    A proof for theorem \ref{T_SP_Rec} is provided in the appendix. A proof for theorem \ref{T_WL_Rec} can be constructed in the same fashion.
\end{note}

We now look to each of the principles defined in the previous section and determine which are satisfied by the simple product method and the weakest link method respectively:
\begin{theorem}\label{SP_Principles}
SP satisfies Anonymity, Premising, Strict Argument, Resilience, Argument Death, Antecedent Maximality, Antecedent Neutrality, Antecedent Weakening, Inferential Weakening, Inference Weight Sensitivity, Proportionality and Weakest-Link Limiting. SP does not satisfy %Counting or
Weakest Link.
\end{theorem}
\begin{theorem}\label{WL_Principles}
WL satisfies Anonymity, Premising, Strict Argument, Resilience, Argument Death, Antecedent Maximality, Antecedent Neutrality, Weakest Link and Weakest-Link Limiting.\\
WL does not satisfy Antecedent Weakening, Inferential Weakening, Inference Weight Sensitivity or Proportionality.% or Counting.
\end{theorem}

A full formal proof for every element of theorem \ref{SP_Principles} can be found in the appendix. Many of the proofs to support theorem \ref{WL_Principles} are very similar to those used for theorem \ref{SP_Principles}. Because of this, combined with the fact that in section \ref{AM_Properties} we take a closer look at the class WL belongs to and how this class relates to our principles, we take a more informal approach with theorem \ref{WL_Principles} and offer a proof sketch in the appendix instead.
\section{Aggregation Methods}
In the previous section we proposed two methods for assigning intrinsic strength to arguments. We saw that the simple product method satisfies most of the principles proposed in section \ref{Principles}. The weakest link method satisfies significantly fewer of the principles, but nevertheless represents an intuitive concept. We also saw that both methods can be rewritten to not derive the assigned strength directly from the basis of an argument as a whole, but rather to assign arguments that state a premise a strength based on the weight of that premise and to assign arguments that make an inference a strength based on an aggregation of the strengths of the argument's antecedents combined with the weight of the applied inference rule.

In this section we introduce a technique that we may use to come up with new methods for assigning strength, based on the rewriting of our previous methods. We combine two functions, one to aggregate the strengths of the antecedents of the argument we are assessing and the other to combine this aggregate with the weight of the top rule used or inference made by the argument. This structure allows us to easily mix and match these two components to fine-tune the behaviour of our strength assigning method.

\begin{definition}[Aggregation Method]\label{AM}
An aggregation method $\text{M} = ( f,g )$ is a pair of functions $g: \bigcup_{n=0}^{\infty} [0,1]^n \rightarrow [0,1]$ and $f: [0,1] \times [0,1] \rightarrow [0,1]$ such that $g$ is symmetric, used to evaluate the intrinsic strength of an argument $A$ over a $\text{WAT} = ( \text{AS}, \K, s )$ such that, when $\{A_1,\dots,A_n\} = \Ant(A)$:
\begin{itemize}
    \item If $\TopRule(A)$ is defined, $\Str(A) = f(s(\TopRule(A)),g(\Str(A_1),\dots,\Str(A_n)))$.
    \item Else, $\Str(A) = f(s(\Conc(A)),g(\Str(A_1),\dots,\Str(A_n)))$.
\end{itemize}
Here $g$ aggregates the strengths of the antecedents of $A$ and $f$ combines the weight of the inference or premise with the aggregated antecedent strengths.
\end{definition}

This definition of an aggregation method is inspired by \cite{DBLP:conf/atal/AmgoudD19}, where a similar method is used to assign arguments an acceptability degree after attacks by other arguments in a fully weighted argumentation graph.\footnote{In \cite{DBLP:conf/atal/AmgoudD19}, a function is used to aggregate the weight of an attack with the weight of the attacker, a second function is used to aggregate the results of the first function for all attacks and a third function is used to combine this second aggregate with the initial weight of the argument under attack.} 

Using our new definition of an aggregation method, we can construct a method that replicates the behaviour of our simple product method; take $M_{sp} = (f_{\text{prod}},g_{\text{prod}})$ where:
\begin{align*}
    f_{\text{prod}}(x,y) &= x \cdot y\\
    g_{\text{prod}}(x_1,\dots,x_n) &= \prod_{i=1}^{n} x_i
\end{align*}
Similarly we can replicate the weakest link method with $M_{wl} = (f_{\text{min}},g_{\text{min}})$, where:
\begin{align*}
    f_{\text{min}}(x,y) &= \text{min}\{x,y\}\\
    g_{\text{min}}(x_1,\dots,x_n) &= \begin{cases} 1 &\mbox{if } n=0 \\
\text{min}\{x_1,\dots,x_n\} &\mbox{otherwise }\end{cases}
\end{align*}

We may now also recombine the functions we used to construct $M_{sp}$ and $M_{wl}$ to create other aggregation methods. For instance we might make a method $M_{wm} = (f_{\text{prod}},g_{\text{min}})$ that looks at the weakest antecedent of an argument and returns the product of said antecedent's strength and the argument's top rule or premise, which we might call the weakening minimum method.

We want the user of the aggregation methods to be free to choose aggregation functions that suit their use case. Therefore we deliberately left the choice of aggregation functions in the definition of an aggregation method unconstrained beyond the required symmetry of $g$. There are, however, some intuitive properties we want our aggregation methods to satisfy. To ensure they do we introduce the notion of a \textit{well-behaved aggregation method}:

\begin{definition}[Well-Behaved Aggregation Method]\label{Well-Behaved}\\
An aggregation method $\text{M} = ( f,g )$ is considered well-behaved iff:
\begin{enumerate}
    \item $f$ is non-decreasing in both variables whenever neither variable is $0$.
    \item $f(0,x) = f(x,0) = 0$
    \item $f(x,1)=f(1,x)=x$
    %\item $f(x,y) > 0 \text{ whenever } x,y>0$
    \item $g()=1$
    \item $g(x)=x$
    \item $g(x_1,\dots, x_n ,0) = 0$
    \item $g(x_1, \dots, x_n) = g(x_1, \dots, x_n, 1)$
    \item $g(x_1, \dots, x_n, y) \leq g(x_1, \dots, x_n, z) \text{ if } y \leq z$
\end{enumerate}\end{definition}

We considered adding two requirements, being $f(x,y) > 0 \textit{ whenever } x,y>0$ and $g(x_1,\dots,x_n) > 0 \textit{ whenever } x_1,\dots,x_n>0$, that correspond to the Resilience principle to the definition of a well-behaved aggregation method. We finally decided against this, because we feel it should be permissible for an aggregation method to 'kill' an argument if its components are too weak. It should be noted, however, that an aggregation method that kills off arguments in this way is precluded from satisfying Resilience.

It is plain to see how both $M_{sp}$ and $M_{wl}$ satisfy all the requirements presented in definition \ref{Well-Behaved} and, as such, are considered well-behaved aggregation methods.

Note how both $f$-functions we proposed are t-norms and both $g$-functions make use of t-norms \cite{DBLP:books/sp/KlementMP00}. It would seem that when one is looking to create a novel aggregation method, t-norms are a good place to look, at least for functions $f$. For functions $g$ one might be tempted to also look to common aggregate functions such as the mean or median, but using these often results in a non-well-behaved aggregation method (specifically violating Definition \ref{Well-Behaved} Point 7).

What follow are a few more examples of functions we could use to construct an aggregation method. For function $f$ we might also use the Hamacher product $f_{\text{Ham}}$ or the Łukasiewicz t-norm $f_{\text{Łuk}}$, both being t-norms:
\begin{align*}
    f_{\text{Ham}}(x,y) &= \begin{cases} 0 &\mbox{if } x=y=0 \\
    \frac{xy}{x+y-xy} & \mbox{otherwise }\end{cases}\\
    f_{\text{Łuk}}(x,y) &= \text{max}(0, x + y - 1)
\end{align*}
We might base our function $g$ on the same two t-norms, adding special cases for when an argument has no or just one antecedent, giving us $g_{\text{Ham}}$ and $g_{\text{Łuk}}$:
\begin{align*}
    g_{\text{Ham}}(x_1, \dots, x_n) \MoveEqLeft= \begin{cases} 1 &\mbox{if } n=0 \\
    x_1 &\mbox{if } n=1 \\
    x_1 \oplus \dots \oplus x_n &\mbox{otherwise }\end{cases}\\
    \text{where } x_1 \oplus x_2 &= \begin{cases} 0 &\mbox{if } x=y=0 \\
    \frac{xy}{x+y-xy} & \mbox{otherwise }\end{cases}\\
    g_{\text{Łuk}}(x_1, \dots, x_n) \MoveEqLeft= \begin{cases} 1 &\mbox{if } n=0 \\
    x_1 &\mbox{if } n=1 \\
    x_1 \oplus \dots \oplus x_n &\mbox{otherwise }\end{cases}\\
    \text{where } x_1 \oplus x_2 &= \text{max}(0, x + y - 1)\\
\end{align*}
Like the functions making up $M_{sp}$ and $M_{wl}$, the functions $f$ and $g$ proposed here satisfy their respective requirements for an aggregation method based on them to be well-behaved.
\section{Properties of Aggregation Methods}\label{AM_Properties}
When choosing functions to construct an aggregation method, t-norms have the clear benefits of being commutative, monotonic and  associative, as well as having $1$ as their identity element and $0$ as a null-element. These properties make it so that when we pick a t-norm for $f$ and we base our $g$ on a t-norm as we did for $g_{\text{Ham}}$, making sure an empty input results in value $1$ and a single input value $x$ results in output value $x$, we are guaranteed to have a well-behaved aggregation method.

We now look at the principles proposed in section \ref{Principles} to see how they relate to (well-behaved) aggregation methods.

Our first observation is that all aggregation methods satisfy Anonymity. This is because two isomorphic arguments have the same shape and the same weights for the premises and inference rules in the same places and an aggregation method only considers these two factors. A formal proof for this can be found in the appendix.
\begin{theorem}\label{AM_Anonymity}
Any aggregation method satisfies Anonymity.
\end{theorem}

The other principles are not necessarily satisfied by any aggregation method. We see that a well-behaved aggregation method is guaranteed to satisfy seven out of our fifteen principles, including Anonymity. This, however, does not mean the other principles cannot be satisfied by an aggregation method. Take, for instance, the well-behaved aggregation method $M_{sp}$ which, as demonstrated in section \ref{Assigning_Strength}, satisfies five of the non-guaranteed principles.

\begin{theorem}\label{WB_Principles}
Beyond Anonymity, any well-behaved aggregation method is guaranteed to satisfy Premising, Strict Argument, Argument Death, Antecedent Maximality, Antecedent Neutrality and Weakest-Link Limiting.\\
The satisfaction of Resilience, Antecedent Weakening, Inferential Weakening, Inference Weight Sensitivity, Proportionality, %Counting
or Weakest Link is not guaranteed.
\end{theorem}

Once again, many of the proofs for theorem \ref{WB_Principles} closely resemble those presented in section \ref{Assigning_Strength}. Some others are simple counterexamples. Because of this we present only a proof sketch, which can be found in the appendix.
\section{Conclusion}
In the field of argumentation, the focus of much of the research done has been on finding sensible ways of interpreting conflicts between arguments. The fruits of this labour can be found in Dung semantics that accept or reject arguments based on the attack relations between them \cite{DBLP:journals/ai/Dung95} and in the many gradual semantics that assign these arguments acceptability degrees \cite{DBLP:conf/atal/AmgoudD19}. Both of these types of semantics belong to what we call abstract argumentation. That is, they do away with the internal structure of the arguments under consideration and look only at the relations between arguments. While gradual semantics have emerged that can take into consideration the weights of attacks and base weights of arguments when determining arguments' final acceptability degree, no standard had emerged for relating these weights to the structure of an argument.

What we set out to do in writing this paper is to investigate how to create suitable methods for taking the structure of an argument and using it, combined with weights for the premises and inference rules used in the argument, to determine the intrinsic strength of an argument. Throughout this investigation we related the method for assigning strength under inspection to a set of principles corresponding to traits we deemed desirable, or at the least intuitive, in a strength-assigning method.

The fact that we take the weights of premises and inference weights to correspond to a degree of defeasibility, with strict rules and premises having weight $1$ and defeasible rules and premises having weight $<1$, gives rise to a possible criticism to our approach. Some might object to the idea of an argument being 'punished' twice for being defeasible: once by having a lowered intrinsic strength and a second time by being opened up to attacks from other arguments.

The first strength-assigning methods we looked at, the simple product method (SP) and the weakest link method (WL), both take the weights of every occurrence of a premise or inference rule in an argument as input when determining said argument's intrinsic strength. We found SP to satisfy many of our principles. WL satisfied fewer of our principles but nevertheless represents an important concept, in that any argument can only be so strong as its weakest link.

Our definition of an argument closely resembles that used in ASPIC+, the system for structured argumentation introduced by Modgil and Prakken \cite{DBLP:journals/argcom/ModgilP14}. This definition allows for an argument to state a premise or to apply an inference rule to the conclusions of a set of antecedent arguments. We saw how both SP and WL can be equivalently redefined to not take the strengths of the members of the basis of an argument as input, but rather to assign an argument stating a premise an intrinsic strength equal to the weight of said premise and for an argument making an inference to combine the weight of the applied inference rule with an aggregation of the strengths of its antecedent arguments. This observation led to the introduction of aggregation methods.

An aggregation method is a combination of two functions that are used to assign an argument an intrinsic strength; one function $g$ to aggregate antecedent argument strengths and another function $f$ to combine the resulting aggregate with the weight of the last applied inference rule or the stated premise. Aggregation methods allow us to mix and match these two functions to easily create new methods of assigning intrinsic strength to an argument.

To capture what behaviours we wanted aggregation methods to express, we introduced the notion of a well-behaved aggregation method. One compelling result here is that we found that whenever one picks a t-norm for $f$ and bases their $g$ on a t-norm, the resulting aggregation method is guaranteed to be well-behaved. A possible criticism to our definition of an aggregation method is that the way it is used to assign an intrinsic strength to an argument is perhaps needlessly complicated. We define special cases for arguments making an inference and those stating a premise, both using functions $f$ and $g$, but what follows from our notion of a well-behaved aggregation method is that we always want premising arguments to get assigned a strength equal to the weight of the premise they state. It might have been more clear to change the case for premising arguments to reflect this directly.

When relating well-behaved aggregation methods to the principles we defined earlier in the paper, we found that a well-behaved aggregation method is only guaranteed to satisfy half of our principles. This finding might give rise to the criticism that when the methods we ourselves deem well-behaved do not necessarily satisfy our own principles, the principles may be too strict. We, however, believe that this result is acceptable as our principles are intended not as options, but rather as formalisations of options.

We see a bright future for aggregation methods in structured argumentation. Future work we would specifically like to see done is an exploration of the weakening of the proposed principles that are not guaranteed to be satisfied by a well-behaved aggregation method, such that they still express the same ideas, but might be satisfied by all well-behaved aggregation methods. Another avenue for future work we believe to be promising is an exploration of the relationships between the proposed principles; we would, for instance, be interested to see whether certain principles follow from (a combination of) other principles or whether there are other properties, such as the guarantee that any subargument of an argument $A$ is at least as strong as $A$, that hold as a consequence of satisfying certain principles. Finally, we feel interesting prospects lie in the development of aggregation methods tailored to specific use cases, making use of the aggregation method's easily adaptable nature.
\subsubsection*{Acknowledgements}
Special thanks go out to Dragan Doder, for his expert guidance in writing this paper, for introducing us to many of the topics we  discuss and for always being there to discuss new ideas. This paper would not have existed without his help.
\bibliographystyle{splncs04}
\bibliography{bibliography}
\section*{Appendix}
%For a full version of this paper, including proofs and proof sketches, please refer to \url{https://arxiv.org/abs/2108.13996}.
%
\begin{proof}Theorem \ref{T_SP_Rec}, the simple product method works recursively.\leavevmode

We are to prove the definitions in theorem \ref{T_SP_Rec} are equivalent to those in definition \ref{SP}. i.e.
\begin{multline*}
    \forall \text{WAT} = ( \text{AS}, \K, s ) \forall A \text{ over WAT},\\(\TopRule(A) = \text{ undefined} \rightarrow s(\Conc(A)) = \prod_{b\in\B(A)}s(b))\\\wedge (\TopRule(A) \neq \text{undefined} \\\rightarrow s(\TopRule(A)) \cdot \prod_{a \in \Ant(A)} \Str_{sp}(a) = \prod_{b\in\B(A)}s(b))
\end{multline*}\leavevmode

Left side of conjunction (Direct Proof):
\begin{enumerate}
    \item \ProofStep{Let $A$ be an argument over a $\text{WAT} = ( \text{AS}, \K, s )$}{Ass.}
    \item \ProofStep{$\TopRule(A) = \text{undefined}$}{Ass.}
    \item \ProofStep{$A$ states a premise.}{2, Def. \ref{Argument}}
    \item \ProofStep{$\B(A) = [\Conc(A)]$}{3, Def. \ref{Argument}}
    \item \ProofStep{$s(\Conc(A)) = \prod_{b\in\B(A)}s(b)$}{4}
\end{enumerate}

Right side of conjunction (direct proof):
\begin{enumerate}
    \item \ProofStep{Let $A$ be an argument over a $\text{WAT} = ( \text{AS}, \K, s )$}{Ass.}
    \item \ProofStep{$\TopRule(A) \neq \text{undefined}$}{Ass.}
    \item \ProofStep{$A$ makes an inference.}{2, Def. \ref{Argument}}
    \item \ProofStep{$\B(A) = \TopRule(A) \uplus \biguplus_{A\pr \in \Ant(A)}\B(A\pr) $}{3, Def. \ref{Argument}}
    \item \ProofStep{$\prod_{b \in \B(A)}s(b) \\= s(\TopRule(A))\prod_{b \in \biguplus_{A\pr \in \Ant(A)}\B(A\pr)}s(b) \\=  s(\TopRule(A)) \prod_{b \in \B(A_1)}s(b) \cdot \dots \cdot \prod_{b\pr \in \B(A_n)}s(b\pr) \\= s(\TopRule(A)) \cdot \Str_{sp}(A_1) \cdot \dots \cdot \Str_{sp}(A_n) \\= s(\TopRule(A)) \cdot \prod_{a \in \Ant(A)} \Str_{sp}(a)$\\ Where $\{A_1, \dots, A_n\} = \Ant(A)$}{4, Def. \ref{SP}}
\end{enumerate}
\qed\end{proof}
\begin{proof}Theorem \ref{SP_Principles}, SP satisfies Anonymity. To prove:\leavevmode
\begin{gather*}
    \forall A A\pr, \text{ if an isomorphism exists between } A \text{ and } A\pr,\ \Str_{sp}(A) = \Str_{sp}(A\pr)
\end{gather*}
We make use of the fact that arguments stating a premise have no antecedents and the observation that any chain of inferences must start from one or more premising arguments to construct an inductive proof.

Base case. (Direct Proof)
\begin{enumerate}
    \item \ProofStep{Let $A,A\pr$ be arguments over a $\text{ WAT } = ( \text{AS}, \K, s )$}{Ass.}
    \item \ProofStep{$A$ and $A\pr$ state a premise.}{Ass.}
    \item \ProofStep{$A$ and $A\pr$ are isomorphic.}{Ass.}
    \item \ProofStep{$s(\Conc(A)) = s(\Conc(A\pr))$}{2, 3, Def. \ref{Isomorphism}}
    \item \ProofStep{$\B(A) = [\Conc(A)]$}{2, Def. \ref{Argument}}
    \item \ProofStep{$\B(A\pr) = \Conc(A\pr)$}{2, Def. \ref{Argument}}
    \item \ProofStep{$\Str_{sp}(A) = \prod_{b\in\B(A)}s(b)$}{Def. \ref{SP}}
    \item \ProofStep{$\Str_{sp}(A\pr) = \prod_{b\in\B(A\pr)}s(b)$}{Def. \ref{SP}}
    \item \ProofStep{$\Str_{sp}(A) = s(\Conc(A))$}{5, 7}
    \item \ProofStep{$\Str_{sp}(A\pr) = s(\Conc(A\pr))$}{6, 8}
    \item \ProofStep{$\Str_{sp}(A\pr) = s(\Conc(A))$}{4, 10}
    \item \ProofStep{$\Str_{sp}(A) = \Str_{sp}(A\pr)$}{9, 11}
\end{enumerate}

Inductive case. (Direct Proof)
\begin{enumerate}
    \item \ProofStep{$\forall a \in \Ant(A)\forall a\pr, a \text{ and } a\pr \text{ are isomorphic} \rightarrow \Str_{sp}(a) = \Str_{sp}(a\pr)$}{Induction Hypothesis}
    \item \ProofStep{Let $A,A\pr$ be arguments over a $\text{ WAT } = ( \text{AS}, \K, s )$}{Ass.}
    \item \ProofStep{$A$ and $A\pr$ make an inference.}{Ass.}
    \item \ProofStep{$A$ and $A\pr$ are isomorphic.}{Ass.}
    \item \ProofStep{$s(\TopRule(A)) = s(\TopRule(A\pr))$}{3, 4, Def. \ref{Isomorphism}}
    \item \ProofStep{there exists a bijective function $f: \Ant(A) \rightarrow \Ant(A\pr)$\\such that $\forall A^{\prime\prime} \in \Ant(A), f(A^{\prime\prime}) \in \Ant(A\pr) \text{ is an isomorphic image of } A^{\prime\prime}$}{3, 4, Def. \ref{Isomorphism}}
    \item \ProofStep{$\prod_{a \in \Ant(A)}\Str_{sp}(a) = \prod_{a\pr \in \Ant(A\pr)}\Str_{sp}(a\pr)$}{1, 6}
    \item \ProofStep{$\Str_{sp}(A) = s(\TopRule(A)) \cdot \prod_{a \in \Ant(A)}\Str_{sp}(a)$}{3, Def. \ref{T_SP_Rec}}
    \item \ProofStep{$\Str_{sp}(A\pr) \\= s(\TopRule(A\pr)) \cdot \prod_{a\pr \in \Ant(A\pr)}\Str_{sp}(a\pr)$}{3, Def. \ref{T_SP_Rec}}
    \item \ProofStep{$\Str_{sp}(A\pr) \\= s(\TopRule(A)) \cdot \prod_{a\pr \in \Ant(A\pr)}\Str_{sp}(a\pr)$}{5, 9}
    \item \ProofStep{$\Str_{sp}(A\pr) \\= s(\TopRule(A)) \cdot \prod_{a \in \Ant(A)}\Str_{sp}(a)$}{7, 10}
    \item \ProofStep{$\Str_{sp}(A) = \Str_{sp}(A\pr)$}{8, 11}
\end{enumerate}
\qed\end{proof}
\begin{proof}Theorem \ref{SP_Principles}, SP satisfies Premising. To prove: \leavevmode
\begin{multline*}
    \forall \text{WAT} = ( \text{AS}, \K, s ) \forall A \text{ over WAT},\\TopRule(A) = \text{ undefined } \rightarrow \Str_{sp}(A) = s(\Conc(A))
\end{multline*}

(Direct Proof)
\begin{enumerate}
    \item \ProofStep{Let $A$ be an argument over a $\text{ WAT } = ( \text{AS}, \K, s )$}{Ass.}
    \item \ProofStep{$\TopRule(A) = \text{ undefined }$}{Ass.}
    \item \ProofStep{$\B(A) = [\Conc(A)]$}{2, Def. \ref{Argument}}
    \item \ProofStep{$\Str_{sp}(A) = \prod_{b\in\B(A)}s(b)$}{Def. \ref{SP}}
    \item \ProofStep{$\Str_{sp}(A) = s(\Conc(A))$}{3, 4}
\end{enumerate}
\qed\end{proof}
\begin{proof}Theorem \ref{SP_Principles}, SP satisfies Strict Argument. To prove:\leavevmode
\begin{gather*}
    \forall \text{WAT} = ( \text{AS}, \K, s ) \forall A \text{ over WAT},\ \DefB(A) = \emptyset \rightarrow \Str_{sp}(A) = 1
\end{gather*}

(Direct Proof)
\begin{enumerate}
    \item \ProofStep{Let $A$ be an argument over a $\text{ WAT } = ( \text{AS}, \K, s )$}{Ass.}
    \item \ProofStep{$\DefB(A) = \emptyset$}{Ass.}
    \item \ProofStep{$\Str_{sp}(A) = \prod_{b\in\DefB(A)}s(b)$}{Def. \ref{SP}}
    \item \ProofStep{$\Str_{sp}(A) = 1$}{2, 3}
\end{enumerate}
\qed\end{proof}
\begin{proof}Theorem \ref{SP_Principles}, SP satisfies Resilience. To prove:\leavevmode
\begin{gather*}
    \forall \text{WAT} = ( \text{AS}, \K, s ) \forall A \text{ over WAT},\ (\forall b \in \B(A), s(b) > 0) \rightarrow \Str_{sp}(A) > 0
\end{gather*}

(Direct Proof)
\begin{enumerate}
    \item \ProofStep{Let $A$ be an argument over a $\text{ WAT } = ( \text{AS}, \K, s )$}{Ass.}
    \item \ProofStep{$\forall b \in \B(A), s(b) > 0$}{Ass.}
    \item \ProofStep{$\Str_{sp}(A) = \prod_{b\in\B(A)}s(b)$}{Def. \ref{SP}}
    \item \ProofStep{$\Str_{sp}(A) > 0$}{2, 3}
\end{enumerate}
\qed\end{proof}
\begin{proof}Theorem \ref{SP_Principles}, SP satisfies Argument Death. To prove:\leavevmode
\begin{gather*}
    \forall \text{WAT} = ( \text{AS}, \K, s ) \forall A \text{ over WAT},\ (\exists b \in \B(A), s(B) = 0) \rightarrow \Str_{sp}(A) = 0
\end{gather*}

(Direct Proof)
\begin{enumerate}
    \item \ProofStep{Let $A$ be an argument over a $\text{ WAT } = ( \text{AS}, \K, s )$}{Ass.}
    \item \ProofStep{$\exists b \in \B(A), s(b) = 0$}{Ass.}
    \item \ProofStep{$\Str_{sp}(A) = \prod_{b\in\B(A)}s(b)$}{Def. \ref{SP}}
    \item \ProofStep{$\Str_{sp}(A) = 0$}{2, 3}
\end{enumerate}
\qed\end{proof}
\begin{proof}Theorem \ref{SP_Principles}, SP satisfies Antecedent Maximality. To prove:\leavevmode
\begin{multline*}
    \forall \text{WAT} = ( \text{AS}, \K, s ) \forall A \text{ over WAT } \forall A\pr \in \Ant(A),\\\Str_{sp}(A\pr) = 1 \wedge \TopRule(A) \neq \text{ undefined } \rightarrow \Str_{sp}(A) = s(\TopRule(A))
\end{multline*}

(Direct Proof)
\begin{enumerate}
    \item \ProofStep{Let $A$ be an argument over a $\text{ WAT } = ( \text{AS}, \K, s )$}{Ass.}
    \item \ProofStep{$\forall A\pr \in \Ant(A), \Str_{sp}(A\pr) = 1$}{Ass.}
    \item \ProofStep{$\TopRule(A) \neq \text{ undefined }$}{Ass.}
    \item \ProofStep{$\Str_{sp}(A) = s(\TopRule(A)) \cdot \prod_{a \in \Ant(A)}\Str_{sp}(a)$}{Theor. \ref{T_SP_Rec}}
    \item \ProofStep{$\Str_{sp}(A) = s(\TopRule(A)) \cdot 1 = s(\TopRule(A))$}{2, 4}
\end{enumerate}
\qed\end{proof}
\begin{proof}Theorem \ref{SP_Principles}, SP satisfies Antecedent Neutrality. To prove:\leavevmode
\begin{multline*}
    \forall \text{WAT} = ( \text{AS}, \K, s ) \forall A A\pr A^{\prime\prime} \text{ over WAT},\\s(\TopRule(A)) = s(\TopRule(A\pr)) \wedge \Ant(A\pr) = \Ant(A) \cup A^{\prime\prime} \wedge \Str_{sp}(A^{\prime\prime}) = 1\\\rightarrow \Str_{sp}(A) = \Str_{sp}(A\pr)
\end{multline*}

(Direct Proof)
\begin{enumerate}
    \item \ProofStep{Let $A, A\pr, A^{\prime\prime}$ be arguments over a $\text{ WAT } = ( \text{AS}, \K, s )$}{Ass.}
    \item \ProofStep{$s(\TopRule(A)) = s(\TopRule(A\pr))$}{Ass.}
    \item \ProofStep{$\Ant(A\pr) = \Ant(A) \cup A^{\prime\prime}$}{Ass.}
    \item \ProofStep{$\Str_{sp}(A^{\prime\prime}) = 1$}{Ass.}
    \item \ProofStep{$\Str_{sp}(A) = s(\TopRule(A)) \cdot \prod_{a \in \Ant(A)}\Str_{SP}(a)$}{Theor. \ref{T_SP_Rec}}
    \item \ProofStep{$\Str_{sp}(A\pr) \\= s(\TopRule(A\pr)) \cdot \prod_{a \in \Ant(A) \cup A^{\prime\prime}}\Str_{SP}(a) \\= s(\TopRule(A\pr)) \cdot 1\prod_{a \in \Ant(A)}\Str_{SP}(a) \\= s(\TopRule(A\pr)) \cdot \prod_{a \in \Ant(A)}\Str_{SP}(a)$}{3, 4, Theor. \ref{T_SP_Rec}}
    \item \ProofStep{$\Str_{sp}(A) = \Str_{sp}(A\pr)$}{2, 5, 6}
\end{enumerate}
\qed\end{proof}
\begin{proof}Theorem \ref{SP_Principles}, SP satisfies Antecedent Weakening. To prove:\leavevmode
\begin{multline*}
    \forall \text{WAT} = ( \text{AS}, \K, s ) \forall A A\pr A^{\prime\prime} \text{ over WAT},\\s(\TopRule(A)) = s(\TopRule(A\pr)) \wedge \Ant(A\pr) = \Ant(A) \cup A^{\prime\prime}\\\wedge \Str_{sp}(A^{\prime\prime}) < 1 \wedge \Str_{sp}(A) > 0\\\rightarrow \Str_{sp}(A) > \Str_{sp}(A\pr)
\end{multline*}

(Direct Proof)
\begin{enumerate}
    \item \ProofStep{Let $A, A\pr, A^{\prime\prime}$ be arguments over a $\text{ WAT } = ( \text{AS}, \K, s )$}{Ass.}
    \item \ProofStep{$s(\TopRule(A)) = s(\TopRule(A\pr))$}{Ass.}
    \item \ProofStep{$\Ant(A\pr) = \Ant(A) \cup A^{\prime\prime}$}{Ass.}
    \item \ProofStep{$\Str_{sp}(A^{\prime\prime}) < 1$}{Ass.}
    \item \ProofStep{$\Str_{sp}(A) > 0$}{Ass.}
    \item \ProofStep{$\Str_{sp}(A) = s(\TopRule(A)) \cdot \prod_{a \in \Ant(A)}\Str_{SP}(a)$}{Theor. \ref{T_SP_Rec}}
    \item \ProofStep{$s(\TopRule(A)) > 0$}{5, 6}
    \item \ProofStep{$\Str_{sp}(A\pr) = s(\TopRule(A\pr)) \cdot \prod_{a \in \Ant(A\pr)}\Str_{SP}(a)$}{Theor. \ref{T_SP_Rec}}
    \item \ProofStep{$\Str_{sp}(A\pr) = \Str_{sp}(A^{\prime\prime}) \cdot s(\TopRule(A)) \cdot \prod_{a \in \Ant(A)}\Str_{SP}(a)$}{2, 3, 8}
    \item \ProofStep{$\Str_{sp}(A\pr) < s(\TopRule(A)) \cdot \prod_{a \in \Ant(A)}\Str_{SP}(a)$}{6, 10}
    \item \ProofStep{$\Str_{sp}(A) > \Str_{sp}(A\pr)$}{2, 5}
\end{enumerate}
\qed\end{proof}
\begin{proof}Theorem \ref{SP_Principles}, SP satisfies Inferential Weakening. To prove:\leavevmode
\begin{multline*}
    \forall \text{WAT} = ( \text{AS}, \K, s ) \forall A \text{ over WAT},\\TopRule(A) \in \Rd \wedge (\forall A\pr \in \Ant(A), Str(A\pr) > 0)\\\rightarrow \Str_{sp}(A) < \text{min}\{ \Str_{sp}(A\pr) | A\pr \in \Ant(A)\}
\end{multline*}

(Direct Proof)
\begin{enumerate}
    \item \ProofStep{Let $A$ be an argument over a $\text{ WAT } = ( \text{AS}, \K, s )$}{Ass.}
    \item \ProofStep{$\TopRule(A) \in \Rd$}{Ass.}
    \item \ProofStep{$s(\TopRule(A)) < 1$}{2, Def. \ref{WAT}}
    \item \ProofStep{$\forall A\pr \in \Ant(A), \Str_{sp}(A\pr) > 0$}{Ass.}
    \item \ProofStep{$\forall A\pr \in \Ant(A), \Str_{sp}(A\pr) \leq 1$}{Def. \ref{WAT}, \ref{SP}}
    \item \ProofStep{$\Str_{sp}(A) = s(\TopRule(A)) \cdot \prod_{A\pr \in \Ant(A)}\Str_{sp}(A\pr)$}{Theor. \ref{T_SP_Rec}}
    \item \ProofStep{$\Str_{sp}(A) < \text{min}\{ \Str_{sp}(A\pr) | A\pr \in \Ant(A)\}$}{3, 4, 5, 6, Def. product}
\end{enumerate}
\qed\end{proof}
\begin{proof}Theorem \ref{SP_Principles}, SP satisfies Inference Weight Sensitivity. To prove:
\begin{multline*}
    \forall \text{WAT} = ( \text{AS}, \K, s ) \forall A A\pr \text{ over WAT},\\\Ant(A) = \Ant(A\pr) \wedge s(\TopRule(A)) < s(\TopRule(A\pr))\\\wedge (\forall A^{\prime\prime} \in \Ant(A), Str(A^{\prime\prime}) > 0)\\\rightarrow \Str_{sp}(A) < \Str_{sp}(A\pr)
\end{multline*}

(Direct Proof)
\begin{enumerate}
    \item \ProofStep{Let $A, A\pr$ be arguments over a $\text{ WAT } = ( \text{AS}, \K, s )$}{Ass.}
    \item \ProofStep{$\Ant(A) = \Ant(A\pr)$}{Ass.}
    \item \ProofStep{$\forall A^{\prime\prime} \in \Ant(A), \Str_{sp}(A^{\prime\prime}) > 0$}{Ass.}
    \item \ProofStep{$s(\TopRule(A)) < s(\TopRule(A\pr))$}{Ass.}
    \item \ProofStep{$\prod_{A^{\prime\prime} \in \Ant(A)}\Str_{sp}(A^{\prime\prime}) > 0$}{3}
    \item \ProofStep{$\Str_{sp}(A) \\= s(\TopRule(A)) \cdot \prod_{A^{\prime\prime} \in \Ant(A)}\Str_{sp}(A^{\prime\prime})$}{Theor. \ref{T_SP_Rec}}
    \item \ProofStep{$\Str_{sp}(A\pr) \\= s(\TopRule(A\pr)) \cdot \prod_{A^{\prime\prime} \in \Ant(A)}\Str_{sp}(A^{\prime\prime})$}{2, Theor. \ref{T_SP_Rec}}
    \item \ProofStep{$\Str_{sp}(A) < \Str_{sp}(A\pr)$}{4, 5, 6, 7}
\end{enumerate}
\qed\end{proof}
\begin{proof}Theorem \ref{SP_Principles}, SP satisfies Proportionality. To prove:
\begin{multline*}
    \forall \text{WAT} = ( \text{AS}, \K, s ) \forall A A\pr \text{ over WAT},\\s(\TopRule(A)) = s(\TopRule(A\pr)) \\\wedge \text{ there exists an injective function } f: \Ant(A) \rightarrow \Ant(A\pr) \\\text{such that } \forall A^{\prime\prime} \in \Ant(A), \Str_{sp}(A^{\prime\prime}) > \Str_{sp}(f(A^{\prime\prime}))\\\rightarrow \Str_{sp}(A) > \Str_{sp}(A\pr)
\end{multline*}

(Direct Proof)
\begin{enumerate}
    \item \ProofStep{Let $A, A\pr$ be arguments over a $\text{ WAT } = ( \text{AS}, \K, s )$}{Ass.}
    \item \ProofStep{$s(\TopRule(A)) = s(\TopRule(A\pr))=1$}{Ass.}
    \item \ProofStep{There exists an injective function $f: \Ant(A) \rightarrow \Ant(A\pr) \\\text{such that } \forall A^{\prime\prime} \in \Ant(A), \Str_{sp}(A^{\prime\prime}) > \Str_{sp}(f(A^{\prime\prime}))$.}{Ass.}
    \item \ProofStep{$\Str_{sp}(A) = s(\TopRule(A)) \cdot \Str_{sp}(A^{\prime\prime}_1) \cdot \dots \cdot \Str_{sp}(A^{\prime\prime}_n)$,\\where $\{A^{\prime\prime}_1) \dots A^{\prime\prime}_n\} = \Ant(A)$}{Theor. \ref{T_SP_Rec}}
    \item \ProofStep{$\Str_{sp}(A\pr) \\= s(\TopRule(A\pr)) \cdot \Str_{sp}(f(A^{\prime\prime}_1)) \cdot \dots \cdot \Str_{sp}(f(A^{\prime\prime}_n)) \cdot \Str_{sp}(A^{\prime\prime\prime}_1) \cdot \dots \cdot \Str_{sp}(A^{\prime\prime\prime}_m)$,\\where $\{f(A^{\prime\prime}_1) \dots f(A^{\prime\prime}_n)\}$ is the image of $\Ant(A)$ under $f$\\and $\{A^{\prime\prime\prime}_1) \dots A^{\prime\prime\prime}_m\} = \Ant(A) \setminus \{f(A^{\prime\prime}_1) \dots f(A^{\prime\prime}_n)\}$}{3, Theor. \ref{T_SP_Rec}}
    \item \ProofStep{$\forall A^{\prime\prime\prime\prime} \in \Ant(A) \cup \Ant(A\pr), 0 \leq \Str_{sp}(A^{\prime\prime\prime\prime}) \leq 1$}{Def. \ref{WAT}, \ref{Basis}, \ref{SP}}
    \item \ProofStep{$\Str_{sp}(A^{\prime\prime}_1) \cdot \dots \cdot \Str_{sp}(A^{\prime\prime}_n) > \Str_{sp}(f(A^{\prime\prime}_1)) \cdot \dots \cdot \Str_{sp}(f(A^{\prime\prime}_n))$}{3, 6}
    \item \ProofStep{$\Str_{sp}(f(A^{\prime\prime}_1)) \cdot \dots \cdot \Str_{sp}(f(A^{\prime\prime}_n)) \cdot \Str_{sp}(A^{\prime\prime\prime}_1) \cdot \dots \cdot \Str_{sp}(A^{\prime\prime\prime}_m)\\\leq \Str_{sp}(f(A^{\prime\prime}_1)) \cdot \dots \cdot \Str_{sp}(f(A^{\prime\prime}_n))$}{5, 6}
    \item \ProofStep{$\Str_{sp}(A^{\prime\prime}_1) \cdot \dots \cdot \Str_{sp}(A^{\prime\prime}_n)\\> \Str_{sp}(f(A^{\prime\prime}_1)) \cdot \dots \cdot \Str_{sp}(f(A^{\prime\prime}_n)) \cdot \Str_{sp}(A^{\prime\prime\prime}_1) \cdot \dots \cdot \Str_{sp}(A^{\prime\prime\prime}_m)$}{7, 8}
    \item \ProofStep{$s(\TopRule(A)) \cdot \Str_{sp}(A^{\prime\prime}_1) \cdot \dots \cdot \Str_{sp}(A^{\prime\prime}_n) \\> s(\TopRule(A\pr)) \cdot \Str_{sp}(f(A^{\prime\prime}_1)) \cdot \dots \cdot \Str_{sp}(f(A^{\prime\prime}_n)) \cdot \Str_{sp}(A^{\prime\prime\prime}_1) \cdot \dots \cdot \Str_{sp}(A^{\prime\prime\prime}_m)$}{2, 9}
    \item \ProofStep{$\Str_{sp}(A) > \Str_{sp}(A\pr)$}{4, 5, 10}
\end{enumerate}
\qed\end{proof}
\begin{proof}Theorem \ref{SP_Principles}, SP does not satisfy Weakest Link. To prove:
\begin{gather*}
    \neg \forall \text{WAT} = ( \text{AS}, \K, s ) \forall A \text{ over WAT},\ \Str_{sp}(A) = \text{min}(\{s(b)| b\in \B(A)\})
\end{gather*}

(Counterexample)
\begin{enumerate}
    \item \ProofStep{Let $A$ be an argument over a $\text{ WAT } = ( \text{AS}, \K, s )$}{Ass.}
    \item \ProofStep{$\Prem(A) = \{p_1\}$}{Ass.}
    \item \ProofStep{$\DefRules(A) = \{d_1\}$}{Ass.}
    \item \ProofStep{$\StrRules(A) = \emptyset$}{Ass.}
    \item \ProofStep{$\B(A) = \{p_1, d_1\}$}{2, 3, 4}
    \item \ProofStep{$s(p_1) = \frac{1}{2}$}{Ass.}
    \item \ProofStep{$s(d_1) = \frac{1}{4}$}{Ass.}
    \item \ProofStep{$\text{min}(\{s(b)| b\in \B(A)\}) = \frac{1}{4}$}{5, 6, 7}
    \item \ProofStep{$\Str_{sp}A = \frac{1}{8}$}{5, 6, 7, Def. \ref{SP}}
    \item \ProofStep{$\Str_{sp}A \neq \text{min}(\{s(b)| b\in \B(A)\})$}{8, 9}
\end{enumerate}
\qed\end{proof}
\begin{proof}Theorem \ref{SP_Principles}, SP satisfies Weakest-Link Limiting. To prove:
\begin{gather*}
    \forall \text{WAT} = ( \text{AS}, \K, s ) \forall A \text{ over WAT},\ \Str_{sp}(A) \leq \text{min}(\{s(b)| b\in \B(A)\})
\end{gather*}

(Direct Proof)
\begin{enumerate}
    \item \ProofStep{Let $A$ be an argument over a $\text{ WAT } = ( \text{AS}, \K, s )$}{Ass.}
    \item \ProofStep{Let $W = \{s(b) | b \in \B(A)\}$}{Ass.}
    \item \ProofStep{Let $m = \text{min}(W)$}{Ass.}
    \item \ProofStep{$\Str_{sp}(A) = \prod_{b\in\B(A)}s(b)$}{Def. \ref{SP}}
    \item \ProofStep{$\Str_{sp}(A) = \prod_{w \in W}w$}{2, 4}
    \item \ProofStep{$\Str_{sp}(A) = m \cdot \prod_{w \in (W \setminus \{m\})}w$}{5}
    \item \ProofStep{$\forall b \in \B(A), s(b) \leq 1$}{Def. \ref{WAT}, \ref{Argument}. \ref{Basis}}
    \item \ProofStep{$\prod_{w \in (W \setminus \{m\})}w \leq 1$}{2, 3, 7}
    \item \ProofStep{$m \cdot \prod_{w \in (W \setminus \{m\})} \leq m$}{8}
    \item \ProofStep{$\Str_{sp}(A) \leq \text{min}(\{s(b)| b\in \B(A)\})$}{2, 3, 6, 9}
\end{enumerate}
\qed\end{proof}
\begin{proof}Theorem \ref{WL_Principles}, proof sketch:
\begin{enumerate}
    \item WL satisfies Anonymity, because it looks only at weights, which are the same across an isomorphism.
    \item WL satisfies Premising, because a premising argument has only one element in its basis, that element being the premise the argument states.
    \item WL satisfies Strict Argument, because when all elements in the basis of an argument have weight $1$, the minimum of those weights is also $1$.
    \item WL satisfies Resilience, because when all elements in the basis of an argument have a non-zero weight, the minimum of those weights is also non-zero.
    \item WL satisfies Argument Death, because the lowest weight possibly assigned to an element of the basis of an argument is $0$. This means that when an element of the basis of an argument has weight $0$, the minimum of the weights of all elements of the basis is also $0$.
    \item WL satisfies Antecedent Maximality, as when all antecedents have the maximum possible weight-value as their strength WL will always assign the weight of the top rule as the strength of the argument.
    \item WL satisfies Antecedent Neutrality. This is the case because $1$ is the highest possible weight for an element in the basis of an argument and consequently the highest possible strength of an argument, so introducing a one-strength antecedent will never be strictly weaker than any of the antecedents or the inference rule already in place.
    \item WL does not satisfy Antecedent Weakening. Take for instance an argument with a top rule with weight $0.2$ and only antecedents with weight $1$, this argument gets assigned strength $0.2$, introducing an argument with strength $0.8$ would not change the minimum and as such the assigned strength even though it is defeasible.
    \item WL does not satisfy Inferential Weakening. Take an argument with top-rule  weight $0.8$ and a single antecedent with strength $0.2$. WL would assign this argument strength $0.2$, equal to (so not strictly lower than) the strength of the weakest antecedent.
    \item WL does not satisfy Inference Weight Sensitivity. Take an argument with a single antecedent with strength $0.2$ and top-rule weight $0.8$. This argument would receive strength $0.2$. The same argument with the top-rule weight lowered to $0.5$ would still be assigned strength $0.2$, no lower than before.
    \item WL does not satisfy Proportionality. Take two argument with top-rule weight $0.2$. Let the one argument have a single antecedent with strength $0.5$ and the other have a single antecedent with strength $0.8$. Both arguments are assigned strength $0.2$ even though both arguments have the same top-rule weight and we can injectively map the antecedents of the second argument to those of the first, such that each of the latter argument's antecedents is strictly stronger than its image.
    % \item WL does not satisfy Counting. Take two arguments with top-rule weight $0.2$. Let the first argument have a single antecedent with strength $1.0$ and the second have a single antecedent with strength $0.8$. Both arguments are assigned strength $0.2$ even though the first has no non-strict antecedents while the second does.
    \item WL satisfies Weakest Link. It prescribes the exact function used by WL to assign strength.
    \item WL satisfies Weakest-Link Limiting as a result of satisfying Weakest Link.
\end{enumerate}
\qed\end{proof}
\begin{proof}Theorem \ref{AM_Anonymity}, any aggregation method satisfies Anonymity. To prove:\leavevmode
\begin{gather*}
    \forall A A\pr, \text{ if an isomorphism exists between } A \text{ and } A\pr,\ \Str_{am}(A) = \Str_{am}(A\pr)
\end{gather*}

We note that, as per definition \ref{Isomorphism}, there are two cases: the $A$ and $A\pr$ both state a premise, or they both make an inference. We also note that any chain of inferences starts with an argument making an inference with only premising arguments as antecedents. We use these facts to construct an inductive proof with an argument stating a premise as the base case and an argument making an inference as the inductive case.

Base case. (direct proof) 
\begin{enumerate}
    \item \ProofStep{Let $A, A\pr$ be arguments over a $\text{ WAT } = ( \text{AS}, \K, s )$}{Ass.}
    \item \ProofStep{Let $\text{am} = (f,g)$ be an aggregation method.}{Ass.}
    \item \ProofStep{An isomorphism exists between $A$ and $A\pr$.}{Ass.}
    \item \ProofStep{$A$ and $A\pr$ state a premise}{Ass.}
    \item \ProofStep{$s(\Conc(A)) = s(\Conc(A\pr))$}{3, 4, Def. \ref{Isomorphism}}
    \item \ProofStep{$\Ant(A) = \Ant(A\pr) = \emptyset$}{4, Def. \ref{Argument}}
    \item \ProofStep{$\TopRule(A) = \TopRule(A\pr)=\text{undefined}$}{4, Def. \ref{Argument}}
    \item \ProofStep{$\Str_{am}(A) \\= f(s(\Conc(A)),g(\Str_{am}(A_1),\dots,\Str_{am}(A_n)))$,\\where $\{A_1,\dots,A_n\} = \Ant(A)$}{2, 7, Def. \ref{AM}}
    \item \ProofStep{$\Str_{am}(A\pr) \\= f(s(\Conc(A\pr)),g(\Str_{am}(A_1),\dots,\Str_{am}(A_n)))$,\\where $\{A_1,\dots,A_n\} = \Ant(A\pr)$}{2, 7, Def. \ref{AM}}
    \item \ProofStep{$\Str_{am}(A) = f(s(\Conc(A)),g())$}{6, Def. \ref{AM}}
    \item \ProofStep{$\Str_{am}(A\pr) = f(s(\Conc(A\pr)),g())$}{6, Def. \ref{AM}}
    \item \ProofStep{$\Str_{am}(A\pr) = f(s(\Conc(A)),g())$}{5, 11}
    \item \ProofStep{$\Str_{am}(A) = \Str_{am}(A\pr)$}{10, 12}
\end{enumerate}

Inductive case. (direct proof)
\begin{enumerate}
    \item \ProofStep{Let $A, A\pr$ be arguments over a $\text{ WAT } = ( \text{AS}, \K, s )$}{Ass.}
    \item \ProofStep{Let $\text{am} = (f,g)$ be an aggregation method.}{Ass.}
    \item \ProofStep{An isomorphism exists between $A$ and $A\pr$.}{Ass.}
    \item \ProofStep{$A$ and $A\pr$ do not state a premise}{Ass.}
    \item \ProofStep{$A$ and $A\pr$ make an inference}{4, Def. \ref{Argument}}
    \item \ProofStep{$s(\TopRule(A)) = s(\TopRule(A\pr))$}{3, 5, Def. \ref{Isomorphism}}
    \item \ProofStep{there exists a bijective function \\$f: \Ant(A) \rightarrow \Ant{A\pr}$ such that \\$\forall a \in \Ant(A), f(a) \in \Ant(A\pr) \\\text{ is an isomorphic image of } a$}{3, 5, Def. \ref{Isomorphism}}
    \item \ProofStep{$\forall a \in \Ant(A), f(a) = b \rightarrow \Str_{am}(a) = \Str_{am}(b)$}{Induction Hypothesis}
    \item \ProofStep{$g(\Str_{am}(a_1),\dots,\Str_{am}(a_n)) \\= g(\Str_{am}(a_1\pr),\dots,\Str_{am}(a_n\pr))$,\\where $\{a_1,\dots,a_n\} = \Ant(A)$, \\$\{a_1\pr,\dots,a_n\pr\} = \Ant(A\pr)$}{2, 7, 8}
    \item \ProofStep{$\TopRule(A) \neq \text{undefined}$, \\$\TopRule(A\pr) \neq \text{undefined}$}{5, Def. \ref{Argument}}
    \item \ProofStep{$\Str_{am}(A) \\= f(s(\TopRule(A)),g(\Str_{am}(a_1),\dots,\Str_{am}(a_n)))$,\\where $\{a_1,\dots,a_n\} = \Ant(A)$}{2, 10, Def. \ref{AM}}
    \item \ProofStep{$\Str_{am}(A\pr) \\= f(s(\TopRule(A\pr)),g(\Str_{am}(a\pr_1),\dots,\Str_{am}(a\pr_n)))$,\\where $\{a\pr_1,\dots,a\pr_n\} = \Ant(A\pr)$}{2, 10, Def. \ref{AM}}
    \item \ProofStep{$\Str_{am}(A) = \Str_{am}(A\pr)$}{6, 9, 11, 12}
\end{enumerate}
\qed\end{proof}
\begin{proof}Theorem \ref{WB_Principles}, proof sketch:
\begin{enumerate}
    \item Any well-behaved aggregation method $M$ is guaranteed to satisfy Premising, as when $\TopRule(A) = \text{undefined}$, $\Ant(A) = \emptyset$ and as such $\Str_M(A) = f(s(\Conc),g())$. Definition \ref{Well-Behaved}, points three and four make it so this equates to $\Str_M(A) = s(\Conc)$.
    \item Any well-behaved aggregation method $M$ is guaranteed to satisfy Strict Argument. All premises being strict means all premising arguments get assigned strength $1$ because a well-defined aggregation method satisfies Premising. This can be used as the base case in an inductive proof where definition \ref{Well-Behaved} points 3, 5 and 7 are used to show that if Strict Argument holds for all antecedent arguments, it is also satisfied by the constructed argument.
    \item A well-behaved aggregation method is not guaranteed to satisfy Resilience as shown by a counterexample. Take well-behaved aggregation method $M_{\text{Łuk}} = ( f_{\text{Łuk}},  g_{\text{Łuk}})$ and an example argument $A$ applying an inference rule with weight $\frac{1}{2}$ to the conclusion of a single antecedent argument with strength $\frac{1}{2}$. Here $\Str_{M_{\text{Łuk}}}(A) = \text{max}(0,\frac{1}{2} + \frac{1}{2} - 1) = 0 \ngtr 0$.
    \item Any well-behaved aggregation method is guaranteed to satisfy Argument Death. This clearly follows from definition \ref{Well-Behaved}, points two and six.
    \item All well-behaved aggregation methods $M$ satisfy Antecedent Maximality. All antecedents having strength $1$ means the result of $g$ is also $1$, as ensured by definition \ref{Well-Behaved}, points five and seven. From here definition \ref{Well-Behaved} point three ensures that $\Str_M(A) = s(\TopRule(A))$.
    \item Any well-behaved aggregation method $M$ satisfies Antecedent Neutrality as it corresponds with definition \ref{Well-Behaved} point 7.
    \item Not all well-behaved aggregation methods satisfy Antecedent Weakening. One of the well-behaved aggregation methods that does not is $M_{wl}$, so the same counterexample can be used as is proposed in the proof sketch for theorem \ref{WL_Principles}.
    \item Not all well-behaved aggregation methods satisfy Inferential Weakening. Again a counterexample in the proof sketch for theorem \ref{WL_Principles} can be used to demonstrate this.
    \item The fact that an aggregation method is well-behaved does not guarantee it satisfies Inference Weight Sensitivity. Once again this is demonstrated by a counterexample in the proof sketch for theorem \ref{WL_Principles}.
    \item Proportionality is also not necessarily satisfied by a well-behaved aggregation method. As above this can be seen from a counterexample presented in the proof sketch for theorem \ref{WL_Principles}.
    % \item Not all well-behaved aggregation methods satisfy Counting as SP, corresponding to a well behaved aggregation method, does not. This is shown in the proof for theorem \ref{SP_Principles}.
    \item Well-behaved aggregation methods are not guaranteed to satisfy Weakest Link, this too is shown in the proof for theorem \ref{SP_Principles}.
    \item All well-behaved aggregation methods satisfy Weakest-Link Limiting. To demonstrate this we use an inductive proof where the base case holds because all well-behaved aggregation methods satisfy Premising. The inductive case can then be proven using definition \ref{Well-Behaved} points 1 and 8 and the knowledge that the weights of premises and inference rules can be no higher than $1$.
\end{enumerate}
\qed\end{proof}
\end{document}